\pdfoutput=1

\documentclass[11pt, table]{article}

\usepackage[final]{acl}
\usepackage{booktabs}
\usepackage{times}
\usepackage{latexsym}
\usepackage{multirow}
\usepackage{rotating}

\usepackage[T1]{fontenc}

\usepackage[utf8]{inputenc}

\usepackage{microtype}

\usepackage{inconsolata}

\usepackage{graphicx}
\usepackage{amsmath}
\usepackage{array}
\usepackage{amsfonts}
\usepackage{xcolor}
\usepackage{pgf}
\usepackage{collcell}
\usepackage{float}
\usepackage{subcaption}
\usepackage{enumitem}
\usepackage{xspace}

\newcommand{\methodname}{\textsc{infini-gram mini}}
\newcommand{\Methodname}{\textsc{Infini-gram mini}}

\newcommand{\huggingface}{\raisebox{-1.5pt}{\includegraphics[height=1.05em]{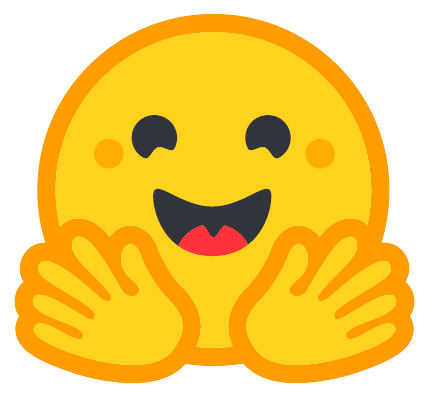}}\xspace}
\newcommand{\internet}{\raisebox{-1.5pt}{\includegraphics[height=1.05em]{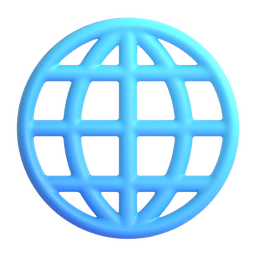}}\xspace}

\newcommand{\github}{\raisebox{-1.5pt}{\includegraphics[height=1.05em]{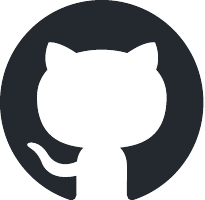}}\xspace}
\newcommand{\home}{\raisebox{-1.5pt}{\includegraphics[height=1.05em]{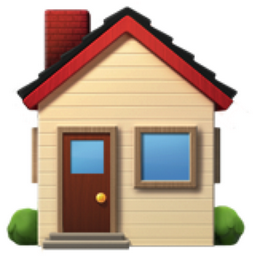}}\xspace}
\newcommand{\aws}{\raisebox{-1.5pt}{\includegraphics[height=1.05em]{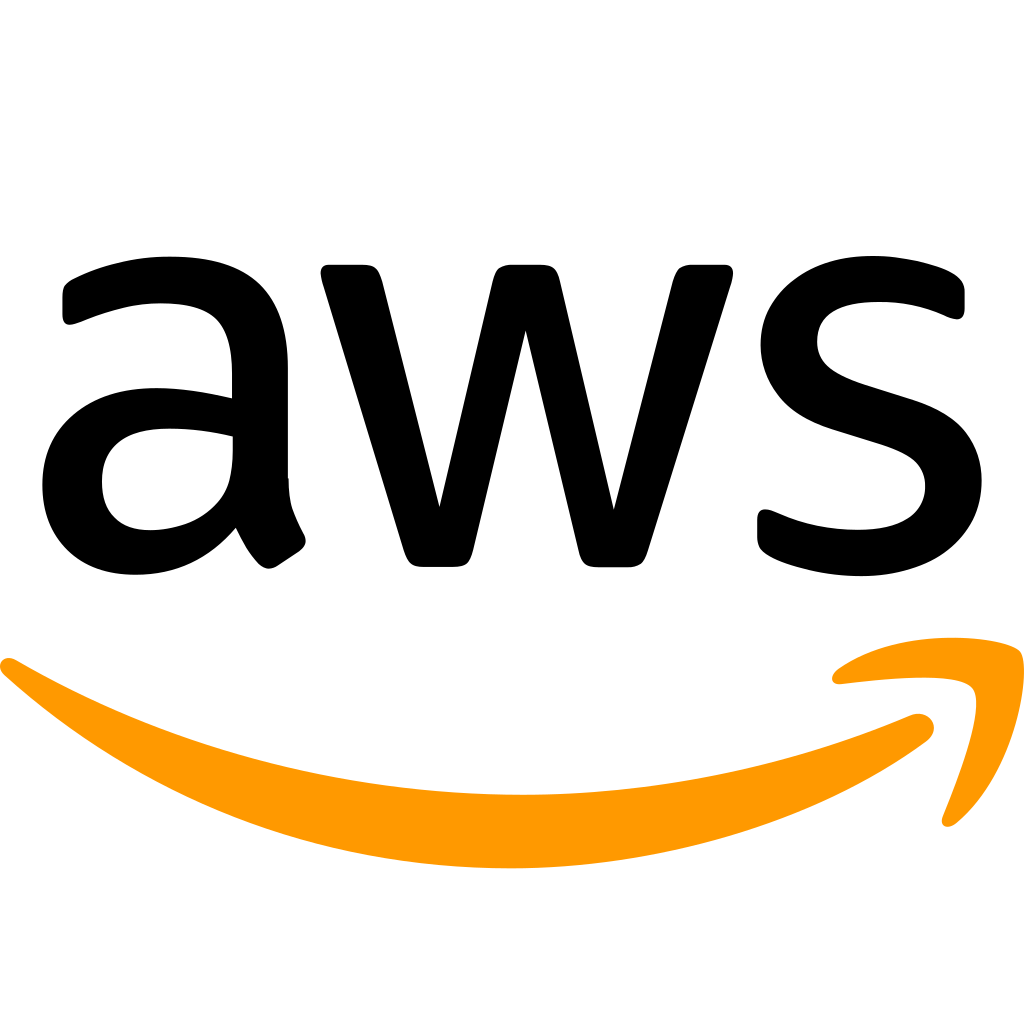}}\xspace}

\definecolor{high}{HTML}{ec462e}
\definecolor{low}{HTML}{ffffff}
\newcommand*{\opacity}{90}
\newcommand*{\MinNumber}{0.0}
\newcommand*{\MidNumber}{7.0}
\newcommand*{\MaxNumber}{15.0}

\newcommand{\colorcell}[1]{
    \pgfmathsetmacro{\value}{#1}

    \pgfmathparse{\value > \MaxNumber ? 1 : 0}
    \ifnum\pgfmathresult=1
        \hspace{-0.33em}\cellcolor{high!\opacity}{#1}
    \else
        \pgfmathparse{\value > \MidNumber ? 1 : 0}
        \ifnum\pgfmathresult=1
            \pgfmathparse{int(round(100*(\value - \MidNumber)/(\MaxNumber - \MidNumber)))}
            \xdef\tempa{\pgfmathresult}
            \hspace{-0.33em}\cellcolor{high!\tempa!yellow!\opacity}{#1}
        \else
            \pgfmathparse{int(round(100*(\MidNumber - \value)/(\MidNumber - \MinNumber)))}
            \xdef\tempa{\pgfmathresult}
            \hspace{-0.33em}\cellcolor{low!\tempa!yellow!\opacity}{#1}
        \fi
    \fi
}

\definecolor{darkblue}{rgb}{0, 0, 0.5}
\hypersetup{colorlinks=true, citecolor=darkblue, linkcolor=darkblue, urlcolor=darkblue}

\title{\Methodname{}: 
Exact n-gram Search at the Internet Scale \\ with FM-Index
}

\newcommand{\aspace}{\hspace{0.7em}}
\newcommand{\uw}{$^{\heartsuit}$}
\newcommand{\aitwo}{$^{\spadesuit}$}
\newcommand{\stanford}{$^{\diamondsuit}$}
\author{
    Hao Xu\uw \aspace
    Jiacheng Liu\uw \aitwo \aspace
    Yejin Choi\stanford \aspace
    Noah A. Smith\uw \aitwo \aspace
    Hannaneh Hajishirzi\uw \aitwo \\
    \uw{}Paul G. Allen School of Computer Science \& Engineering, University of Washington \\
    \aitwo{}Allen Institute for AI \aspace
    \stanford{}Stanford University
}

\begin{document}

\maketitle

\begin{abstract}
Language models are trained mainly on massive text data from the Internet, and it becomes increasingly important to understand this data source. 
Exact-match search engines 
enable searching in large text corpora -- counting string appearances and retrieving the enclosing documents -- yet the high storage overhead hinders their application on Internet-scale data.
We present \textbf{\methodname{}}, an efficient and scalable system that can make petabyte-level
text corpora searchable. 
Based on the FM-index data structure \citep{892127}, which simultaneously indexes and compresses text, 
our system creates indexes with size only 44\% of the corpus.
\Methodname{} greatly improves upon the best existing implementation of FM-index in terms of indexing speed (18$\times$) and memory use during both indexing (3.2$\times$ reduction) and querying (down to a negligible amount). 
We index 83TB of Internet text in 99 days with a single CPU node with 128 vCPUs (or 19 hours if using 137 such nodes).
We show one important use case of \methodname{} in a large-scale analysis of benchmark contamination.
We find several core LM evaluation benchmarks to be heavily contaminated in Internet crawls (up to 74.2\% in GSM8K),
which could lead to overestimating the capabilities of language models if trained on such data.
We host a benchmark contamination bulletin to share the contamination rate of many core and community-contributed benchmarks.
We also release a web interface and an API endpoint to serve general search queries on \methodname{} indexes.
\end{abstract}

\renewcommand{\arraystretch}{1.2}
\begin{table}[!b]
    \small
    \setlength{\tabcolsep}{2pt}
    \resizebox{\linewidth}{!}{
    \begin{tabular}{ccl}
         \home & \textbf{Project Home} & \href{https://infini-gram-mini.io/}{\path{infini-gram-mini.io}} \\
         \huggingface & \textbf{Web Interface} & \href{https://infini-gram-mini.io/demo}{\path{infini-gram-mini.io/demo}} \\
         \aws & \textbf{API Endpoint} & \href{https://api.infini-gram-mini.io/}{\path{api.infini-gram-mini.io}} \\
         \internet & \textbf{Documentation} & \href{https://infini-gram-mini.io/docs}{\path{infini-gram-mini.io/docs}} \\
         \github & \textbf{Source Code} & \href{https://infini-gram-mini.io/code}{\path{infini-gram-mini.io/code}} \\
         \huggingface & \textbf{Contam Bulletin} & \href{https://infini-gram-mini.io/bulletin}{\path{infini-gram-mini.io/bulletin}}
    \end{tabular}
    }
\end{table}
\renewcommand{\arraystretch}{1}

\begin{figure}[!t]
    \centering
    \includegraphics[width=0.95\linewidth]{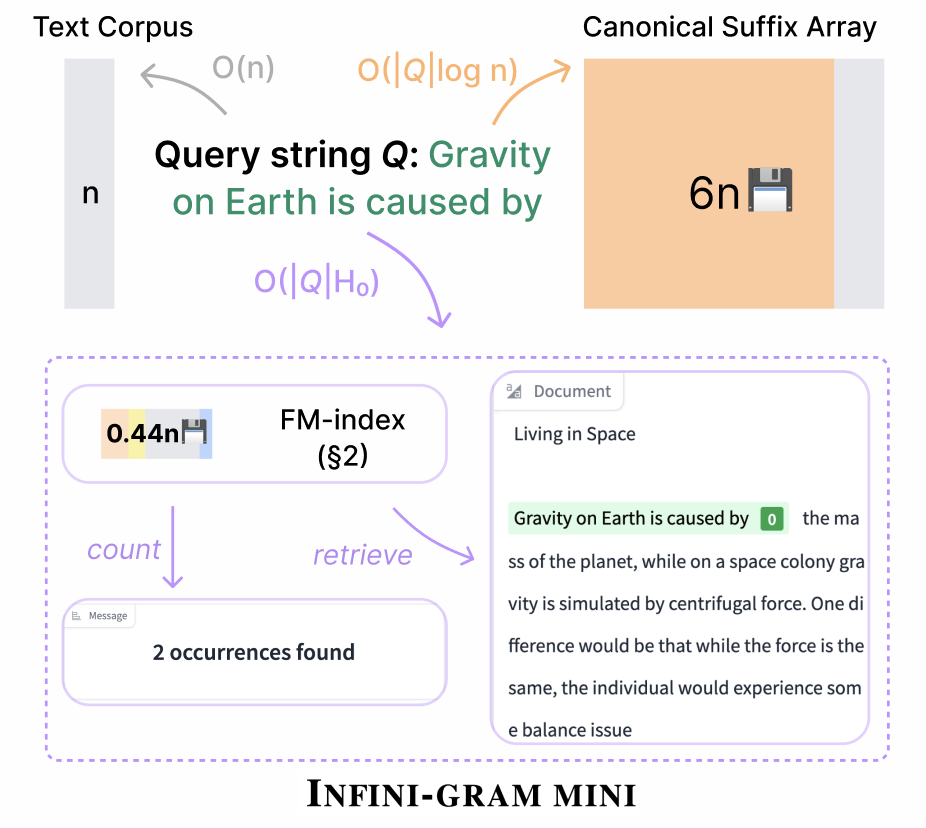}
    \caption{
    Overview of \methodname{}. Based on the FM-index data structure, \methodname{} supports efficient exact-match search in massive text corpora 
    ($n \simeq 10^{15}$ bytes)
    while reducing the index size down to 7\% compared to a canonical suffix array index. 
    Searching naively in the corpus would have time complexity of $O(n)$
    and is thus impractical; with \methodname{}, the search time complexity is independent of $n$. 
    $|Q|$ is the length of query string and can be arbitrarily long, and $H_0 \approx 2.1$ is the zeroth-order entropy of the text corpus. 
    } 
    \label{fig:overview}
\end{figure}

\section{Introduction}

Modern language models (LMs) are trained mainly on text datasets downsampled from massive, petabyte-level text corpora like Common Crawl \citep{commoncrawl}. As these LMs are deployed more broadly, it becomes more pressing to understand the training data and its effects on model behavior \citep{Liu2025OLMoTraceTL,Han2022ORCAIP}. 
As a starting point, we want to make these text corpora \emph{searchable}; in particular, searching for \emph{exact} matches of \emph{long} sequences has gained increasing interest \citep{elazar2024whatsbigdata,Merrill2024EvaluatingNN,Lu2024AIAH}.
The size of these corpora makes this problem extremely challenging, creating demand for more efficient indexing techniques. 

Prior systems for exact-match search build indexes several times as large as the text corpora.
\citet{Merrill2024EvaluatingNN} index 1.3TB of text with a suffix automaton (storage multiplier 29$\times$); \citet{Liu2024InfiniGram} index 12TB of text with a suffix array (storage multiplier 6$\times$); \citet{elazar2024whatsbigdata} index 35TB of text with the proprietary ElasticSearch engine (storage multiplier about $2\times$).
The size of these indexes renders it impractical to apply them to petabyte-level corpora.

To address this challenge, we index text corpora with \textbf{FM-index} \citep{892127}, a compact data structure frequently used in bioinformatics \citep{guo2025effloc, Depuydt_Renders_Abeel_Fostier_2023, wang-genomic, 10.1093/bioinformatics/btu541}, 
but not yet used for natural language data at scale.
Prior work only applies this data structure to 13.4 GB dataset \citep{bevilacqua2022autoregressivesearchenginesgenerating}. 
We explain FM-index in detail in \S\ref{sec:background}.
For natural text, the size of FM-index can be made as small as 0.26$\times$ of the corpus;
practically, to ensure low query latency, we build FM-index with a storage multiplier of \textbf{0.44$\times$}, or 7\% compared to the canonical suffix array. 

\textbf{\Methodname{}} is the system we developed for efficiently constructing the FM-index at scale and answering search queries. 
For indexing, we extend and combine components from \citet{Liu2024InfiniGram}, \citet{sdsl}, and \citet{LABEIT20172} into a highly parallelized program, which achieves a \textbf{18$\times$} speedup and uses only \textbf{32\%} as much RAM compared to the best existing implementation by \textsc{SDSL} \citep{sdsl}.\footnote{\href{https://github.com/simongog/sdsl-lite}{\path{https://github.com/simongog/sdsl-lite}}} 
We use \methodname{} to construct FM-index for two LM pretraining datasets -- the Pile and DCLM-baseline -- and 7 recent Common Crawl dumps from January to July 2025.
Altogether, we indexed \textbf{83TB} of text in 99 days with a single CPU node with 128 vCPUs, and this could have been done in 19 hours if embarrassingly parallelized across 137 such nodes. 
We estimate that indexing the full Common Crawl would take 1,200 node-days, or 19 hours if parallelized across 1,500 nodes.
For answering queries, we extend \textsc{SDSL} to work with on-disk index, reducing the RAM requirement to a negligible amount.
Our inference engine supports counting the occurrences of a query string and retrieving documents that contain the query string, both within seconds when working on the above corpora.


We apply \methodname{} to \textbf{analyzing contamination of benchmarks} widely used in state-of-art LM evaluations (\S\ref{sec:analyzing-contamination}).
\Methodname{} allows us to do the analysis on larger corpora than prior works and on new benchmarks uploaded. This would be much more expensive if using other indexing methods.
We find several core benchmarks to be heavily contaminated (\S\ref{sec:contamination-results}).
\Methodname{} detects exact overlap for question of contaminated entries in text corpora, among which a large majority of questions appear together with the correct answer.
\textbf{This reveals a dire evaluation crisis:} as benchmarks get increasingly contaminated by Internet crawls and consequently LM training data, evaluation results may give inflated estimates of true model capabilities.
As such, we host a benchmark contamination bulletin to continually monitor contamination of core and community-contributed benchmarks on new Internet snapshots, and we call for more community attention on this matter.


Beyond contamination analysis, \methodname{} opens up more impactful use cases such as task-specific dataset construction and pretraining data curation.
We release a \textbf{web interface} and \textbf{API endpoint} of \methodname{}, so that everyone can search in the text corpora that we have indexed.
We plan to continue indexing new corpora and share regular updates. We also release our source code. We hope this tool can enable more insightful analysis and better use of massive text corpora.  


\section{Background: FM-index}
\label{sec:background}

The FM-index \citep{892127} is a full-text index that supports efficient pattern matching, counting, and text retrieval on a highly compressed representation of the text corpus. 
Compared with a canonical suffix array, FM-index stores a compact variation of a suffix array and the text corpus, greatly reducing storage overhead.

\subsection{Data Structures and Implementation}
\label{sec:data_structure}

\begin{figure}[!t]
    \centering
    \includegraphics[width=1\linewidth]{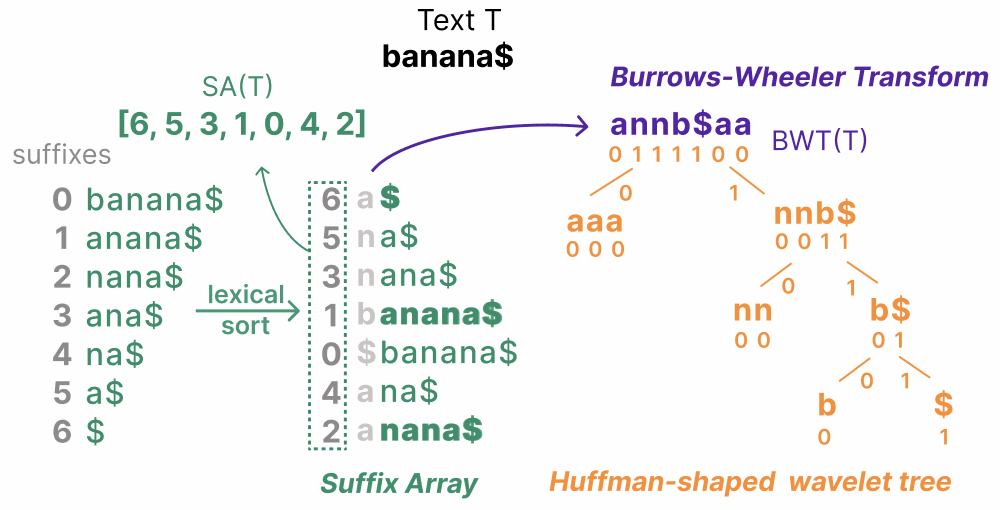}
    \caption{
    The FM-index data structure (\S\ref{sec:data_structure}) used in \methodname{}, shown for a toy string with length $n = 7$.
    The \textcolor{teal}{suffix array} is sampled with a sampling rate $a=3$ and only elements corresponding to \textbf{\textcolor{teal}{bolded}} suffixes are stored.
    The \textcolor{violet}{BWT} can be derived from the SA, and is stored in compressed form as a \textcolor{orange}{Huffman-shaped wavelet tree.} 
    }
    \label{fig:fm-index}
\end{figure}

On a high level, FM-index has two core components: (1) a sampled suffix array and its inverse, and (2) the Burrows-Wheeler transform represented using a Huffman-shaped wavelet tree. Below we describe each component as applied to a single string (appropriate for our application; see \S\ref{sec:method}). \autoref{fig:fm-index} shows a toy example illustrating the data structure.

\paragraph{Suffix array and sampling.} 
The suffix array $\mathit{SA}$ of a string $T$ of length $n$ is an array of integers representing the starting positions of all suffixes of $T$ in lexicographic order. Formally, $\mathit{SA}[i]=j$ if the suffix $T[j...n]$ ranks $i$th among all suffixes in lexicographical ordering. 
The suffix array enables quickly locating the positions where a pattern appears in the string. 
Canonically, storing the whole suffix array would take $O(n \log n)$ space, or $5n$ bytes for strings with length up to one trillion \citep{Liu2024InfiniGram}. To reduce storage overhead, FM-index samples $\mathit{SA}$ at regular intervals, storing only every $a$th entry where $a$ is a parameter. 
When querying the index, if we need to access an unsampled $\mathit{SA}$ entry, we can recover its value by referencing the BWT data structure, which is introduced below.

\paragraph{Burrows-Wheeler transform (BWT).} 
BWT \citep{Burrows1994ABL} 
rearranges the string $T$ (with termination symbol \$) into a reversible permutation $L$ that clusters repeated symbols. The BWT is defined as a string $L$ where $L[i]=T[\mathit{SA}[i]-1]$ if $SA[i]>0$, or $L[i]=\$$ otherwise. Intuitively, $L$ is the concatenation of the symbol preceding each suffix when the suffixes are sorted in the SA order (see \autoref{fig:fm-index}). 

A key property of BWT is the Last-to-First (LF) mapping, which maps the $i$th occurrence of a symbol $c$ in $L$ to the $i$th occurrence of it in $F$, where $F[i]=T[\mathit{SA[i]}]$. The LF mapping is defined as $\mathit{LF}(i)=C[c]+\text{rank}(c, i)$, where $C[c]$ is the number of symbols in $L$ lexicographically smaller than $c$, and $\text{rank}(c,i)$ counts the occurrences of $c$ in $L[0...i]$. This property allows us to traverse the string in reverse order, which is essential for finding patterns in and reconstructing (part of) the string $T$ from the index.


\paragraph{Huffman-shaped wavelet trees.}
Given the BWT $L$ and the sampled SA, we can reconstruct the original string $T$ from these data structures so we don't need to store $T$. However, $L$ still takes as much storage as $T$. To further compress $L$, a wavelet tree is used to represent $L$ by hierarchically partitioning the alphabet.
In the wavelet tree, each leaf node represents a symbol in the alphabet, and each non-leaf node stores a bitvector marking whether the symbol at each position of $L$
belongs to the left or right subtree.
A Huffman-shaped wavelet tree \citep{Makine2005huff} optimizes the hierarchy by grouping and coding symbols based on their frequencies in $L$.
Storing it costs $(nH_0+2\sigma\log n)$ bits, where $\sigma$ is the alphabet size, and $H_0$ is the zeroth-order entropy\footnote{zeroth-order entropy measures the unigram probability distribution of individual symbols (i.e. UTF-8 characters).} of $L$ and $H_0 \approx 0.26 \log_2{\sigma}$ in our experiments with natural language text; this is smaller than storing $L$ directly, which is $n \log_2{\sigma}$ bits.
Besides compression, the tree can efficiently support two operations crucial to LF mapping: $\text{rank}(c, \ell)$ counts the number of occurrences of a character $c$ in $L[0...\ell-1]$, and $\text{select}(c, m)$ finds the position of the $m$th occurrence of $c$ in $L$.
Both operations have a time complexity of $O(H_0)$.

\paragraph{Inverse suffix array.}
To allow reconstructing part of the original string $T$ from any given position, FM-index also stores a sampled version of the inverse suffix array $\mathit{ISA}$.
$\mathit{ISA}$ is the inverse of the permutation specified in $\mathit{SA}$ and is defined as $\mathit{ISA}[j]=i$ if $\mathit{SA}[i]=j$. 
When reconstructing $T$ from a given position, $\mathit{ISA}$ is used to identify the corresponding rank in the suffix array to start reconstruction.
In FM-index, we can use a different sampling rate for $\mathit{ISA}$, storing only every $b$th entry, where $b$ can be larger
than $a$ as $\mathit{ISA}$ is used less.

\subsection{Querying the FM-index}
\label{sec:operations}


There are
three basic operations supported by FM-index: \textit{find}, \textit{locate}, and \textit{reconstruct}.
Counting a pattern can be done with a find operation, and retrieving segments containing the pattern from the original string can be done with a combination of all three operations. See App.~\S\ref{app:fm-query} for more details.

\section{\Methodname{}}
\label{sec:method}

We develop \methodname{} as a scalable and efficient implementation of FM-index on natural language data.
For indexing, \methodname{} achieves 18$\times$ speedup and reduces RAM usage by 3.2$\times$ compared to the previous best existing implementation in \textsc{SDSL}.
This paves way for indexing petabyte-level text corpora with reasonable time and compute. 
For querying, we extended the implementation in \textsc{SDSL} to work with indexes stored on disk, allowing us to query large indexes on machines with limited RAM.


In contrast to the original infini-gram which tokenizes the text, we construct the index directly on raw UTF-8 bytes,
which saves tokenization time and allows more flexibility when querying.
Tokenization is a technique to reduce index size, but since FM-index is already a compressed index, tokenization would not be helpful here.


\subsection{Index Construction}



\paragraph{Optimizing the indexing steps.}
We use a parallelized implementation of each indexing step.
For building SA and BWT, we adapt from the parallel implementation in \citet{Liu2024InfiniGram}.
For building the wavelet tree, we use the parallel implementation in \citet{LABEIT20172}.
For sampling SA and creating a sampled version of ISA, we parallelized the implementation in \textsc{SDSL}.
All these steps were single-threaded in \textsc{SDSL}, and by parallelizing them we achieve significant speedup.

We measured the time and peak RAM usage for indexing an 8.7 GB corpus (a single file from DCLM-baseline) using implementation of \textsc{SDSL} and \methodname{}. Their implementation required 5,847 seconds and 74,807 MB of peak RAM, whereas \methodname{} completed indexing in 324 seconds (18$\times$ speedup) and peak RAM of 23,742 MB (3.2$\times$ reduction).

\paragraph{Preprocessing.}
FM-index is designed to work on a single string.
To index a text corpora, which consists of a collection of documents, we encode all documents with UTF-8 and concatenate all these byte arrays with the \texttt{\textbackslash xff} byte (not used by UTF-8) to mark document boundaries.
We then construct FM-index for this big byte string.
We work on UTF-8 bytes rather than characters to keep the alphabet size small ($\sigma = 256$).
Along with this \textit{text index}, we also store a \textit{text offset} file that records the starting position of each document, which is useful for retrieving whole document and metadata.
Aside from the actual text, we also index the metadata of the documents to make the metadata searchable while storing it in a similar compressed form.

\paragraph{Partitioning.}
For large corpora, we partition it into multiple shards and index each shard independently. Searching across multiple shards produces the same result as searching a single, larger index, and this allows us to build indexes with limited RAM per node as well as embarrassingly parallelize across multiple nodes. 

\paragraph{Indexed corpora.}
We have built the index for the following corpora: the Pile (1.3TB training set, 1.4GB validation set; \citealp{pile}), DCLM-baseline (17TB; \citealp{li2024dclm}), and the Common Crawl between January and July 2025 (
``CC-2025-05'' to ``CC-2025-30''
; 7 crawls and 65TB total; \citealp{commoncrawl}).\footnote{We extract text from the CC crawls with resiliparse, following \citet{li2024dclm}.}

\begin{table*}
    \centering
    \resizebox{0.85\textwidth}{!}{%
    \begin{tabular}{lcccc}
        \toprule
        \textbf{Dataset} & \textbf{Original Size } & \textbf{Indexing Time } & \textbf{ Index Size} & \textbf{Num. Shards}\\
        & (TB) & (CPU node-days) & (TB) & \\
        \midrule
        Pile-validation & 0.0001345        & 0.00057 & 0.000602 (0.45$\times$) & 1 \\
        Pile-train & 1.308      &  1.31  & 0.588 (0.45$\times$) & 2 \\
        DCLM-baseline & 16.666 & 12.6   & 7.523 (0.45$\times$) & 25 \\
        CC-2025-05 & 9.079 &  11.8  & 3.972 (0.44$\times$) & 15 \\
        CC-2025-08 & 8.163 &  10.6  & 3.574 (0.44$\times$) & 15 \\
        CC-2025-13 & 10.393 & 13.6 & 4.563 (0.44$\times$) & 17 \\
        CC-2025-18 & 10.498 & 13.7 & 4.664 (0.44$\times$) & 17 \\
        CC-2025-21 & 9.221 & 12.1 & 4.302 (0.46$\times$) & 15 \\
        CC-2025-26 & 8.724 & 11.4 & 3.835 (0.44$\times$) & 15 \\
        CC-2025-30 & 9.006 & 11.7 & 3.952 (0.44$\times$) & 15 \\
        \midrule
        \textbf{Total} & \textbf{83.059} & \textbf{98.8} & \textbf{36.884 (0.44$\times$)} & 137 \\
        \bottomrule
    \end{tabular}
    }%
    \caption{Text corpora we indexed with \methodname{}, along with indexing time and index size. The reported numbers only include actual document content and do not include metadata.}
    \label{construct-table}
\end{table*}

\paragraph{Indexing time.}
We use CPU nodes with 128 vCPUs and 2TiB RAM to construct indexes, and under this constraint, each shard can be as large as 700GB which can be indexed in 12--19 hours.
\autoref{construct-table} reports the time for indexing the above corpora.
Indexing time shows a slightly super-linear increase with respect to shard size, and we report detailed stepwise breakdown in App.~\S\ref{app:latency-breakdown}. 
If we were to index the full Common Crawl dataset (about 1PB), we would split it into 1,500 shards, which can be indexed in 1,200 node-days.

\paragraph{Index size.}
We choose the sampling rate empirically to balance storage savings and query latency. In our implementation, we sample every $a=32$ entries of SA and every $b=64$ entries of ISA.
This yields indexes with 0.44$\times$ the size of the corpora (\autoref{construct-table}).
Conceptually, if we set $a$ and $b$ to be very big, then the SA and ISA would have negligible size, and the index can be as small as 0.26$\times$ the size of the corpus.

\subsection{Querying with \methodname{}}

Similar to other exact-match search engines, \methodname{} supports two types of query: \textit{counting} the number of occurrences of a string, and \textit{retrieving documents} that contain a string.
At query time, \methodname{} keeps all index files on disk as memory-mapped files, thus requiring only minimal CPU and RAM (loading all indexes uses only $\sim30$ MB if RAM and 1 vCPU in our setting).

Querying in \methodname{} is slower and more complex than in canonical suffix arrays due to the compressive nature of FM-index.
The SA is subsampled, and the original text is shuffled and compressed, both requiring additional random disk reads to recover.
The latency of queries is largely determined by the disk I/O performance.

\paragraph{Counting a string.}
We use FM-index's \textit{find} operation to count the number of occurrences of a query string.
We parallelize this operation across all $k$ shards, and report the sum of counts across all shards.
The total number of random disk reads is $O(k |Q| H_0)$, where $|Q|$ is the length of the query string.
If the disk's I/O throughput is high enough, the query latency can be as low as $O(|Q| H_0)$. 

\paragraph{Retrieving documents.}
We first use the \textit{find} operation to get the range in the SA that corresponds to the query string.
Each element in this range indicates one appearance of the query string in the text corpus.
If the index has multiple shards, we will have one range for each shard.
For each element in this range, we can use the \textit{locate} operation to find the position of the match in the original text $T$, then get the boundaries of the enclosing document with a binary search in the text offset file, and finally use the \textit{reconstruct} operation to get the document text.
The number of random disk reads for \textit{locate} is $O(a H_0)$, and for \textit{reconstruct} is $O((b + d) H_0)$ where $d$ is the document length.
We parallelize \textit{reconstruct} operation by dividing document text into up to $t=10$ chunks, or chunks of length 100 if $d<1000$.
Retrieving a single document is parallelized across $t$ threads, and it can be further parallelized for retrieving multiple documents.

\paragraph{Query latency.}
\Methodname{} presents second-level latency on both types of query. We benchmark the query latency with the index files stored on Google Cloud Platform (GCP) SSD disks with 80,000 IOPS and 1200 MB/s throughput. 
We measure the average latency over 100 queries for each query type and setting.
For counting, short queries ($|Q| \le 10$) can be handled within 0.4 seconds for all corpora we indexed, while longer queries ($|Q|=1000$) are handled in 8 seconds for CC-2025-05 and 25 seconds for DCLM-baseline.
The difference in query latency is caused by the different number of shards in the corpora.
For document retrieval, retrieving a snippet of $d=100$ bytes can be done within 2 seconds for all corpora, and retrieving $d=3000$ bytes takes up to 4.5 seconds.
See App.~\S\ref{app:inference-benchmarking} for benchmarking details.

\subsection{Web Interface and API Endpoint}

We host a 
Hugging Face 
interface for easy access to counting 
and document retrieval 
with 83 TB corpora
(App.~\S\ref{app:interfce}).
We also release an API endpoint that allows programmatically submitting requests.

\section{Analyzing and Monitoring Benchmark Contamination with \methodname{}}
\label{sec:analyzing-contamination}

In this section, we showcase how \methodname{} can be used to analyze benchmark contamination at scale. \Methodname{} allows us to search in the largest body of text ever in the open-source community with minimal storage overhead, enabling large-scale benchmark contamination analysis at low cost. Our work also supports analyzing new benchmarks uploaded later.

 By identifying lexical overlap between 24
widely-used evaluation benchmarks and three major corpora, we find non-trivial contamination across 12 benchmarks (\S\ref{sec:contamination-results}).
We then retrieve documents accounting for contamination for further analysis (\S\ref{sec:contamination-analysis}).
Powered by \methodname{}, we develop a monitoring system with a benchmark contamination bulletin, where we continuously monitor new crawls and report benchmarks' contamination status over time (\S\ref{sec:bulletin}). We also invite the community to contribute by suggesting additional benchmarks or uploading new ones for analysis.

\Methodname{} has further potential applications where exact-match search is needed, including (1) task-specific dataset construction, where creation of those datasets requires retrieving documents containing specific terms or phrases, and (2) data curation, where \methodname{} can assist in identifying and removing duplicate, low-quality, or sensitive text and documents. 
We leave these directions for future exploration.

\subsection{Setup}
\paragraph{Contamination detection method.}
We check contamination of test examples by measuring lexical overlap with the text corpus, following standard practice in the literature \citep{brown2020languagemodelsfewshotlearners, grattafiori2024llama3herdmodels}.
Given a text entry, we extract all 50-character substrings $S$ with a stride of one word, and determine entry contamination based on the proportion of substrings that appear at least once in the corpus:
$$\eta = \frac{\sum_{s \in S} \mathbb{I} \big[ \texttt{count}(s)>0 \big] }{|S|}.$$

We classify an entry into three contamination levels using the same thresholds with different namings 
as \citet{Touvron2023Llama2O}:
\begin{itemize}[leftmargin=12pt, itemsep=-4pt, topsep=4pt]
    \item Clean, if $\eta < 20\%$
    \item Suspicious, if $20\% \leq \eta < 80\%$
    \item Dirty, if $\eta \geq 80\%$
\end{itemize}

\paragraph{Benchmarks.}
To comprehensively evaluate contamination, we analyze a broad range of benchmarks widely used to evaluate state-of-the-art LMs. We categorize benchmarks into five groups based on their primary focus: (1) {\textbf{knowledge and reasoning}}, i.e., understanding and reasoning over factual and conceptual knowledge; (2) {\textbf{math}}, i.e., mathematical reasoning ability; (3) {\textbf{code}}   generation and modification abilities; (4) {\textbf{commonsense understanding}}, i.e., commonsense knowledge and reasoning abilities; and (5) {\textbf{reading comprehension}}  of textual context.
In total, we analyze 24 benchmarks; see App.~\S\ref{app:benchmark-citation} for the full list and citations.

For benchmarks with question-answering format, we specifically check the question entries for contamination. Questions are usually longer and contain sufficient contextual information to uniquely identify the benchmark example, whereas answers can be short, such as a multiple choice option or a number. For language understanding tasks that involve reading a context or paragraph before answering, we check the context entries for contamination. 

We evaluate on only the test set of benchmarks. For benchmarks with large test set, we downsample to 1,000 entries for efficiency. For benchmarks with multiple subtasks, we sample proportionally from each subtask to maintain representative distribution. 

\subsection{Results}
\label{sec:contamination-results}
\autoref{tab:benchmark_contamination} reports the percentage of dirty entries in benchmarks against the Pile (knowledge cutoff in 2020), DCLM-baseline (knowledge cutoff in 2022), and seven CC crawls (knowledge cutoff in January-July 2025).
The detailed count of dirty and suspicious entries can be found in App.~\S\ref{app:more-contamination-result}.


\paragraph{Many widely-used benchmarks are highly contaminated.} When checking against DCLM-baseline, MMLU has 27.70\% dirty entries, MMLU-Pro has 16.20\%, ARC-Challenge has 32.6\%, and ARC-Easy has 32.3\%. These benchmarks, especially MMLU, have been used to evaluate virtually every new LM in recent years. 
We believe that the observed contamination levels are a strong signal that many recently reported results may overestimate language model abilities on truly new, unseen evaluation items.

\paragraph{Larger and newer corpora show greater contamination.} Compared with Pile-train, most benchmarks show a higher  dirty rate on DCLM-baseline. For example, ARC-Challenge has 17$\times$ more and MMLU has 1.1$\times$ more dirty entries. However, CC-2025-05 has lower contamination rate than DCLM-baseline on most benchmarks, which is likely because DCLM-baseline corpus is larger and is a high-quality subset of crawls spanning a decade, while CC-2025-05 is an unfiltered single crawl.

\paragraph{Contamination level varies by domain.} Benchmarks in historically-significant domains, such as commonsense understanding (e.g., ARC, OpenbookQA) and reading comprehension (e.g., SQuAD, CoQA), tend to show higher dirty rates on all corpora, while those in emerging domains like math (e.g., GSM8K, MATH-500) and code (e.g., HumanEval, LiveCodeBench) remain relatively clean at this writing.

\paragraph{New benchmarks are gradually getting contaminated in recent corpora.} Newer benchmarks are initially clean on earlier corpora. For example, AIME-2024 is uncontaminated on Pile-train and DCLM-baseline, but show 40.00\% dirty rate on the more recent CC-2025-21; similarly, GPQA shows 2.70\% dirty rate on CC-2025-26. GSM8K is very clean on Pile-train, DCLM-baseline, and earlier CC-2025 crawls, but has 74.2\% dirty entries on CC-2025-21. 

\begin{table*}[!t]
\centering
\resizebox{\textwidth}{!}{%
\begin{tabular}{l rrrrrrrrr}
\toprule
\textbf{} & \textbf{Test} & \textbf{Pile} & \textbf{DCLM} & \textbf{CC} & \textbf{CC} & \textbf{CC} & \textbf{CC} & \textbf{CC} & \textbf{CC} \\
\textbf{} & \textbf{Size} & \textbf{train} & \textbf{baseline} & \textbf{2025-05} & \textbf{2025-08} & \textbf{2025-13} & \textbf{2025-18} & \textbf{2025-21} & \textbf{2025-26} \\
\midrule

\multicolumn{10}{c}{\textbf{Knowledge and Reasoning}} \\
MMLU        & 1000 & \colorcell{13.20} & \colorcell{28.40} & \colorcell{13.50} & \colorcell{9.00} & \colorcell{12.10} & \colorcell{11.50} & \colorcell{11.70} & \colorcell{9.20} \\
MMLU-Pro    & 1000 & \colorcell{5.50}  & \colorcell{16.20} & \colorcell{7.10}  & \colorcell{5.40} & \colorcell{6.00}  & \colorcell{6.30}  & \colorcell{7.40}  & \colorcell{6.90} \\
BigBenchHard& 1000 & \colorcell{0.00}  & \colorcell{0.10}  & \colorcell{1.40}  & \colorcell{1.40} & \colorcell{3.20}  & \colorcell{2.30}  & \colorcell{1.80}  & \colorcell{1.70} \\
AGIEval     & 1000 & \colorcell{0.80}  & \colorcell{3.10}  & \colorcell{2.70}  & \colorcell{3.60} & \colorcell{3.00}  & \colorcell{7.00}  & \colorcell{9.40}  & \colorcell{4.60} \\
GPQA        & 448  & \colorcell{0.00}  & \colorcell{0.00}  & \colorcell{0.90}  & \colorcell{2.00} & \colorcell{1.30}  & \colorcell{0.70}  & \colorcell{0.90}  & \colorcell{2.70} \\
HLE         & 881  & \colorcell{0.00}  & \colorcell{0.30}  & \colorcell{0.10}  & \colorcell{0.00} & \colorcell{0.10}  & \colorcell{0.00}  & \colorcell{0.00}  & \colorcell{0.00} \\

\midrule
\multicolumn{10}{c}{\textbf{Math}} \\
AIME-2024   & 30   & \colorcell{0.00}  & \colorcell{0.00}  & \colorcell{10.00} & \colorcell{3.30} & \colorcell{6.70}  & \colorcell{40.00} & \colorcell{40.00} & \colorcell{13.30} \\
GSM8K       & 1000 & \colorcell{0.00}  & \colorcell{5.00}  & \colorcell{5.00}  & \colorcell{0.80} & \colorcell{6.90}  & \colorcell{0.70}  & \colorcell{74.20} & \colorcell{7.30} \\
MATH-500    & 500  & \colorcell{0.60}  & \colorcell{3.20}  & \colorcell{0.60}  & \colorcell{7.80} & \colorcell{0.80}  & \colorcell{0.80}  & \colorcell{0.80}  & \colorcell{8.20} \\
MGSM        & 250  & \colorcell{0.00}  & \colorcell{0.00}  & \colorcell{5.60}  & \colorcell{1.60} & \colorcell{35.60} & \colorcell{0.80}  & \colorcell{72.80} & \colorcell{6.00} \\

\midrule
\multicolumn{10}{c}{\textbf{Code}} \\
HumanEval   & 164  & \colorcell{0.00}  & \colorcell{0.00}  & \colorcell{0.00}  & \colorcell{0.60} & \colorcell{0.60}  & \colorcell{0.60}  & \colorcell{0.00}  & \colorcell{0.00} \\
HumanEval+  & 164  & \colorcell{0.00}  & \colorcell{0.00}  & \colorcell{0.00}  & \colorcell{0.60} & \colorcell{0.60}  & \colorcell{0.60}  & \colorcell{0.00}  & \colorcell{0.00} \\
LiveCodeBench&880  & \colorcell{0.00}  & \colorcell{0.00}  & \colorcell{0.00}  & \colorcell{0.00} & \colorcell{0.00}  & \colorcell{0.00}  & \colorcell{0.00}  & \colorcell{0.00} \\
SWE-bench   & 500  & \colorcell{0.00}  & \colorcell{0.00}  & \colorcell{0.20}  & \colorcell{0.20} & \colorcell{0.00}  & \colorcell{0.00}  & \colorcell{0.00}  & \colorcell{0.00} \\
MBPP        & 500  & \colorcell{0.00}  & \colorcell{0.40}  & \colorcell{1.00}  & \colorcell{1.40} & \colorcell{1.20}  & \colorcell{1.80}  & \colorcell{1.00}  & \colorcell{1.40} \\

\midrule
\multicolumn{10}{c}{\textbf{Commonsense Understanding}} \\
ARC-Challenge&1000 & \colorcell{1.80}  & \colorcell{34.10} & \colorcell{11.90} & \colorcell{4.00} & \colorcell{3.10}  & \colorcell{3.80}  & \colorcell{4.20}  & \colorcell{4.80} \\
ARC-Easy    & 1000 & \colorcell{1.30}  & \colorcell{31.70} & \colorcell{5.40}  & \colorcell{9.50} & \colorcell{5.50}  & \colorcell{5.50}  & \colorcell{6.10}  & \colorcell{6.20} \\
CSQA        & 1000 & \colorcell{0.10}  & \colorcell{1.00}  & \colorcell{0.10}  & \colorcell{0.10} & \colorcell{0.20}  & \colorcell{0.10}  & \colorcell{0.00}  & \colorcell{0.10} \\
HellaSwag   & 1000 & \colorcell{0.00}  & \colorcell{0.00}  & \colorcell{0.00}  & \colorcell{0.00} & \colorcell{0.00}  & \colorcell{0.00}  & \colorcell{0.00}  & \colorcell{0.10} \\
OpenbookQA  & 500  & \colorcell{10.80} & \colorcell{15.60} & \colorcell{14.60} & \colorcell{30.20}& \colorcell{13.20} & \colorcell{13.40} & \colorcell{13.20} & \colorcell{12.20} \\
Social IQa  & 1000 & \colorcell{0.00}  & \colorcell{0.50}  & \colorcell{0.20}  & \colorcell{4.40} & \colorcell{0.20}  & \colorcell{0.30}  & \colorcell{0.20}  & \colorcell{0.10} \\
WinoGrande  & 1000 & \colorcell{0.00}  & \colorcell{0.00}  & \colorcell{0.00}  & \colorcell{0.00} & \colorcell{0.00}  & \colorcell{0.00}  & \colorcell{0.00}  & \colorcell{0.00} \\

\midrule
\multicolumn{10}{c}{\textbf{Reading Comprehension}} \\
CoQA        & 500  & \colorcell{8.00}  & \colorcell{18.40} & \colorcell{7.40}  & \colorcell{8.80} & \colorcell{8.60}  & \colorcell{7.20}  & \colorcell{7.60}  & \colorcell{8.80} \\
SQuAD       & 1000 & \colorcell{2.80}  & \colorcell{40.10} & \colorcell{2.70}  & \colorcell{33.00}& \colorcell{10.10} & \colorcell{1.50}  & \colorcell{2.00}  & \colorcell{8.50} \\

\bottomrule
\end{tabular}%
}
\caption{Dirty rates for benchmarks across the Pile, DCLM-baseline, and Common Crawl from January to July, 2025. Full result is reported in \autoref{tab:full_benchmark}. For benchmark entries with over 1000 entries, we report dirty rate on the downsampled subset. Cell background color indicates benchmark cleaniness, with more redness representing increasing levels of contamination.
}
\label{tab:benchmark_contamination}
\end{table*}

\begin{figure*}
    \centering
    \includegraphics[width=0.95\linewidth]{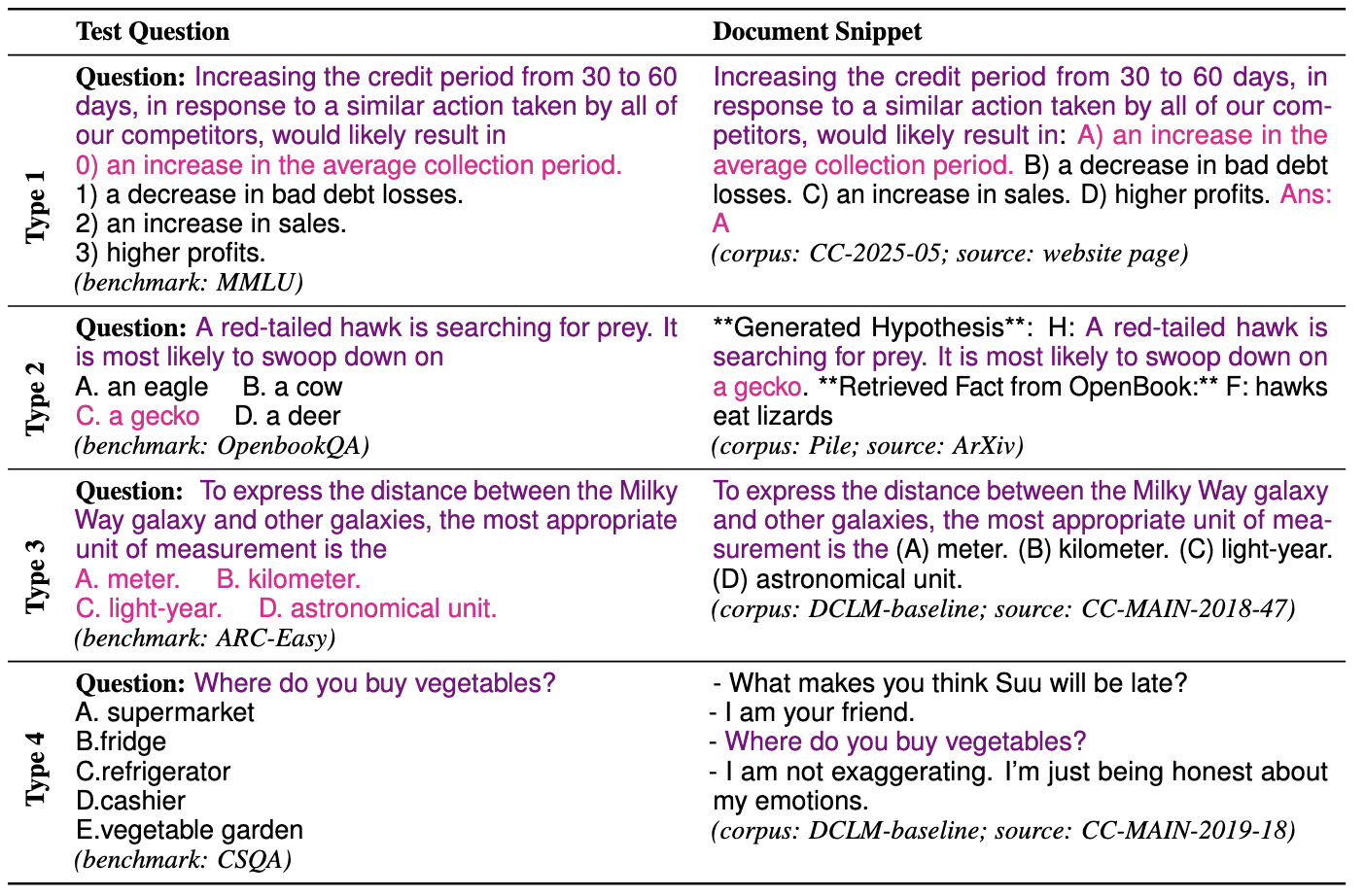}
    \caption{Examples of four contamination types. \textcolor{violet}{\textbf{Violet}} text is the text overlap between benchmark entry and corpus. \textcolor{magenta}{\textbf{Magenta}} text is the mapping of answers.}
    \label{tab:contam-example}
\end{figure*}

\subsection{Analysis}
\label{sec:contamination-analysis}
With \methodname{}, we can retrieve documents from the corpus that contain contaminated examples for further analysis. We use LLM-as-a-judge to categorize all dirty instances in Pile-train, DCLM-baseline, and CC-2025-05 into 
following scenarios (see App.~\S\ref{app:annotation} for more details), with examples shown in Table \ref{tab:contam-example}:

\begin{itemize}[leftmargin=12pt, itemsep=-4pt, topsep=4pt]
    \item \textbf{Type 1: Question and answer appear as-is in corpus.} The question and correct answer appear in the  corpus in the exact format as in benchmark entry. The model may memorize entirely from these data instead of performing capabilities examined in benchmarks. This accounts for 72.5\% of all dirty entries in Pile-train, 82.6\% on DCLM-baseline, and 58\% on CC-2025-05.
    \item \textbf{Type 2: Question appeared, answer in natural language.} The answer is expressed in natural language rather than the exact benchmark format, and model may directly infer the correct answer from it. This accounts for 4.5\% of all dirty entries in Pile-train, 2.1\% on DCLM-baseline, and 6.2\% on CC-2025-05.
    \item \textbf{Type 3: Question appeared, no corresponding answer.} The question (and maybe the multiple-choice options) appeared in the corpus, but correct answer is missing. This accounts for 18.1\% of all dirty entries in Pile-train, 10.9\% on DCLM-baseline, and 30.2\% on CC-2025-05.
    \item \textbf{Type 4: False positives.} There are documents that superficially match the question but are unrelated to the benchmark example. This happens on entries with a very short question. This accounts for 3.1\% of all dirty entries in Pile-train, 1\% on DCLM-baseline, and 3\% on CC-2025-05.
\end{itemize}



We show 7 contamination examples in App.~\S\ref{app:examples} and their source of contamination. We found that contamination is caused by (1) the benchmark is sourced from the Internet (e.g., AIME-2024, \autoref{fig:example_aime}; SWE-bench, \autoref{fig:example_swebench}), (2) there are other benchmarks or online quizzes that are sourced from the benchmark (e.g, GSM8K, \autoref{fig:example_gsm8k}), (3) blogs and papers cite a benchmark entry as example (e.g, GPQA, \autoref{fig:example_gpqa}; BigBenchHard, \autoref{fig:example_bbh}; OpenbookQA, \autoref{fig:example_obqa}), and (4) benchmark entry coincide with online source (e.g., MMLU, \autoref{fig:example_mmlu}).

Contamination could cause LLM to overperform on evaluation benchmarks by enabling models to retrieve memorized answers from training data rather than performing task-specific reasoning. Our finding shows that a large majority of dirty entries contain exact matches of both question and answer, though even subtler forms of contamination such as question-only matches and paraphrased answers could inflate benchmark scores. As training corpora grow, the risk of benchmark contamination and the potential for rote memorization increase, making it more important to  decontaminate training corpora and construct contamination-free benchmarks to avoid overestimating model capabilities.

\subsection{Benchmark Contamination Bulletin} 
\label{sec:bulletin}
Using \methodname{}, we implement a benchmark contamination monitoring system that tracks benchmark contamination. In the future, we will keep indexing the latest crawl in Common Crawl and update contamination results to track benchmark contamination as corpora evolve. 
The system also allows anyone to add or upload new benchmarks to be monitored, fostering collaboration in benchmark monitoring. See App.~\S\ref{app:bulletin-interface} for system interface.

\section{Related Work}
\paragraph{Exact-match search in large text corpora.}
Prior work has used different techniques to enable exact-match full-text search in large text corpora, including suffix arrays, suffix automata, and proprietary search engines.
Below we survey the largest-scale implementation of each technique known to us.
\citet{Merrill2024EvaluatingNN} apply a suffix automaton to 1.3TB of text.
\citet{Liu2024InfiniGram} apply a suffix array to 12TB of text.
\citet{elazar2024whatsbigdata} use the proprietary ElasticSearch to index and analyze 35TB of text.
All these methods have significant storage multiplier wrt the size of text indexed, ranging from ElasticSearch's 2$\times$ to suffix automata's 29$\times$.
In contrast, our FM-index-based index has a storage multiplier as small as 0.26$\times$, allowing us to index the largest body of text ever in the open-source community. 

\paragraph{Benchmark contamination.} 
Benchmark contamination appeared as a critical concern in LLM evaluations in recent studies. 
Prior works has quantified benchmark contamination on open-sourced models using various matching strategies, including n-gram or token overlap \citep{soldaini-etal-2024-dolma, grattafiori2024llama3herdmodels, soldaini-etal-2024-dolma, olmo20252olmo2furious}, longest substring match \citep{singh2024evaluationdatacontaminationllms}, skipgram match \citep{Touvron2023Llama2O}, and full-text exact match \citep{elazar2024whatsbigdata}. \citet{sainz2024datacontaminationreport2024} reports data contamination from multiple sources through shared efforts. However, indexing large-scale pretraining corpora and performing exhaustive searches is computationally expensive, and prior studies on open corpora are limited in their scale (up to RedPajama-1T and Dolma, 12TB).
To the best of our knowledge, our work conducts contamination analysis on the largest open corpora to date.


\section{Conclusion}
We introduce \methodname{}, an efficient system for indexing text with 0.44$\times$ their original size, enabling efficient counting and searching in massive text corpus, and we show its scalability to a petabyte-scale corpus. We showcase \methodname{}'s application on benchmark contamination analysis at scale.

\section*{Limitations}
Although \methodname{} shows significant improvements on text compression rate, document retrieval latency remains higher than a canonical suffix array. For example, reconstructing a 3000-character document takes 1.8 seconds on Pile and 4.5 seconds on DCLM-baseline, compared to millisecond-level retrieval in system like Infini-gram \citep{Liu2024InfiniGram}. This is a trade-off with \methodname{}'s compression rate: \methodname{} does not store the original text in a contiguous block. To retrieve a document, we need to reconstruct it character-by-character, leading to a large number of reads in random addresses. These process takes long time since the entire index is kept on disk at inference time. The high latency can potentially be reduced using techniques like disk page prefetching.

Identifying co-occurrences of multiple patterns is inefficient with the FM-index, since mapping every match from the suffix array range back to its position in original text takes very long time thus making this operation impractical. In contrast, the original infini-gram supports this operation efficiently.

\Methodname{} only supports exact-match searches. As a result, our benchmark contamination analysis is limited to case-sensitive exact matching, which may fail to detect contamination of instances with minor textual discrepancies. 

As the text corpora may contain biased or toxic content, document retrieval output may contain content that can be perceived as ethically problematic, and may contain sensitive information. The output of \methodname{} does not reflect authors' views.


\section*{Acknowledgments}
We thank Christina Boucher (University of Florida) for telling us about the FM-index data structure.
We would like to thank members of H2lab and AllenNLP for sharing their valuable feedback on this project. This work was funded in part by NSF IIS-2044660 and IIS-2113530, and by RS-2024-00457882, National AI Research Lab Project.

\bibliography{main}

\clearpage

\appendix

\section{Querying the FM-Index}
\label{app:fm-query}
This section introduces in details three operations supported by FM-index: \textit{find}, \textit{locate}, and \textit{reconstruct}.

\begin{figure*}[!t]
    \centering
    \includegraphics[width=\linewidth]{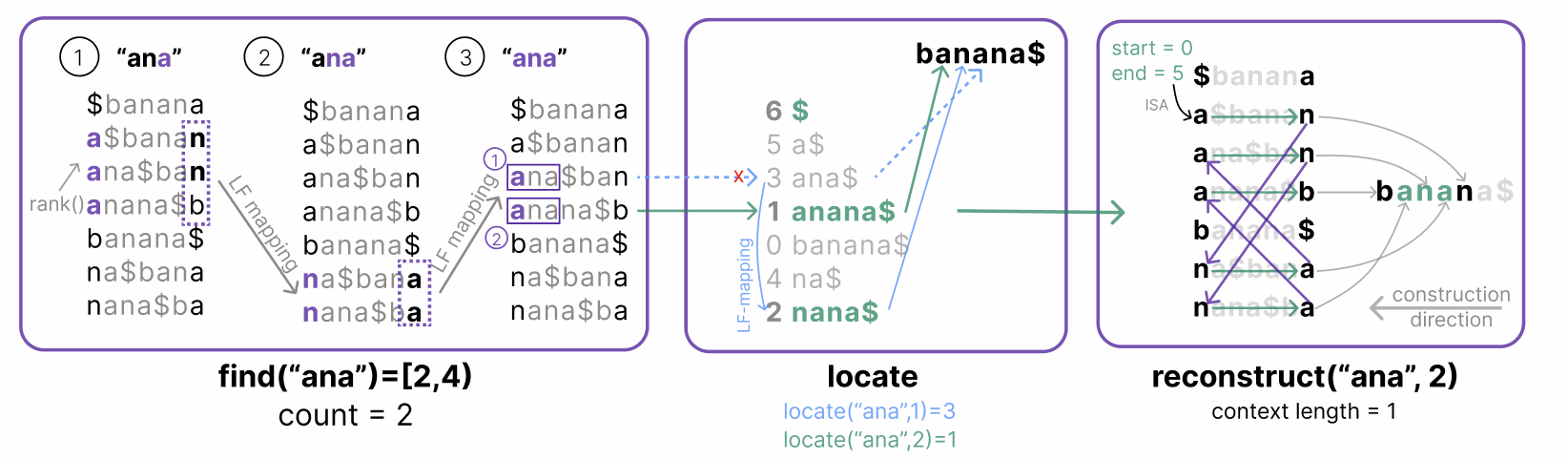}
    \caption{
    Illustration of operations on FM-index (\S\ref{sec:operations}, App.~\S\ref{app:fm-query}). \textbf{Left:} \textit{find} operation computes the SA range corresponding to all occurrences of the pattern. \textbf{Middle:} \textit{locate} operation computes the position of pattern occurrence in the original string for each position in the SA range. \textbf{Right:} \textit{reconstruct} operation gets a substring of the original string enclosing the second pattern occurrence with a context length of 1. The occurrence ranking is based on its order in SA.
    }
    \label{fig:algo}
\end{figure*}

\paragraph{Find.}
Each occurrence of a pattern string $Q$ 
in the haystack string $T$ corresponds to an element in the SA, and all these occurrences live in a consecutive range in the SA.
The \textit{find} operation computes this range with backward search:
starting with the full range, it iterates through the symbols in $Q$ in reverse order, with each iteration narrowing the range using the character table $C$ and the rank operation of the BWT's wavelet tree. The length of the final range is the count of $Q$ in $T$. \autoref{fig:algo} (left) illustrates finding pattern ``ana'' in string ``banana''. The time complexity of \textit{find} is $O(|Q| H_0)$, where $|Q|$ is the length of the pattern.

\paragraph{Locate.}
The \textit{locate} operation maps any position in the SA to its corresponding position in $T$. Since the SA is sampled at interval $a$, for an unsampled position $i$, the algorithm applies LF mapping at most $a$ times to locate the nearest sampled SA entry. \autoref{fig:algo} (middle) illustrates locating the two occurrences of ``ana'' in ``banana''. The time complexity of locating each position is $O(t_{\mathit{SA}})$, where $t_{\mathit{SA}}$ is the complexity of accessing an SA entry, which is $O(1)$ for sampled entries and $O(aH_0)$ for unsampled entries.

\paragraph{Reconstruct.} 
After getting the position of a pattern occurrence in $T$, we can \textit{reconstruct} a substring of $T$ enclosing that occurrence with some additional context. Starting at the end position of the desired substring, we apply the LF mapping to traverse through the BWT and use ISA to recover the symbol, and reconstruct symbols in reverse order until reaching the start position of the desired substring. \autoref{fig:algo} (right) shows reconstructing the second occurrence of ``ana'' with context length of 1. The time complexity of \textit{reconstruct} is $O(d H_0 + t_{\mathit{ISA}})$, where $d$ is the length to reconstruct, and $t_{\mathit{ISA}}$ is the complexity of accessing an ISA entry, which is $O(1)$ for sampled entries and $O(bH_0)$ for unsampled entries.

\section{Indexing Time}
\label{app:latency-breakdown}

\autoref{tab:construction_time_breakdown} presents indexing time for each shard of text corpora with stepwise breakdown.
Suffix array construction scales super-linearly depending on the level of duplication \citep{lee-etal-2022-deduplicating}. Constructing alphabet, wavelet tree, and sampling are linear operations. 
Indexing time also varies across similar-size shards originated from different text corpora, showing the time also depend on text corpora properties. For example, over 60\% duplicate documents in the Pile \citep{elazar2024whatsbigdata} causes SA construction to take significantly longer compared to DCLM-baseline. 

\begin{table*}[h!]
    \centering
    \begin{tabular}{l|rrrr}
        \toprule
        \textbf{Step} & \textbf{Pile-val} & \textbf{Pile-train}  & \textbf{DCLM-baseline} & \textbf{CC-2025-05} \\ 
        & \textbf{(1.4 GB)} & \textbf{(653 GB)} & \textbf{(667 GB)}  & \textbf{(654 GB)} \\  \midrule
        SA+BWT & 29 s & 41710 s  & 29543 s & 55692 s \\ \midrule
        alphabet & 4 s & 2584 s  & 2895 s & 2580 s \\ \midrule
        wavelet tree & 13 s & 5773 s  & 6257 s & 5325 s \\ \midrule
        sample SA & 1 s & 2540 s  & 2232 s & 2013 s \\ \midrule
        sample ISA & 2 s & 3975 s  & 2659 s & 2313 s \\ \midrule
        \textbf{Total} & \textbf{49 s} & \textbf{15.7 h} & \textbf{12.1 h} & \textbf{18.9 h} \\ \bottomrule
    \end{tabular}
    \caption{Index construction time for \textbf{each shard} of the text corpora, with stepwise breakdown. The size of text in each shard is noted at the top. To get the indexing time of the full corpus on a single node, roughly multiply by the number of shards; though this can be embarrassingly parallelized across multiple nodes. Metadata size and metadata indexing time are excluded.}
    \label{tab:construction_time_breakdown}
\end{table*}

\section{Benchmarking Query Latency}
\label{app:inference-benchmarking}

\begin{table*}[!h]
    \centering
    \resizebox{\textwidth}{!}{%
    \begin{tabular}{llcccc}
        \toprule
         & & \textbf{Pile-train} & \textbf{DCLM-baseline} & \textbf{CC-2025-05} & \textbf{Time} \\
        & & ($n$ = 1.3T) & ($n$ = 16.7T) & ($n$ = 9.1T) & \textbf{Complexity} \\
        & & (S = 2) & (S = 25) & (S = 15) \\
        \midrule
        \textbf{Counting a query of length $|Q|$} & & & & & $O(|Q| H_0)$ \\
        \quad \ldots ($|Q|$ = 1) & & 0.004 s & 0.005 s & 0.032 s \\
        \quad \ldots ($|Q|$ = 2) & & 0.015 s & 0.017 s & 0.094 s \\
        \quad \ldots ($|Q|$ = 5) & & 0.061 s & 0.207 s & 0.206 s \\
        \quad \ldots ($|Q|$ = 10) & & 0.106 s & 0.402 s & 0.350 s \\
        \quad \ldots ($|Q|$ = 20) & & 0.182 s & 0.868 s & 0.638 s \\
        \quad \ldots ($|Q|$ = 50) & & 0.393 s & 1.743 s & 1.063 s \\
        \quad \ldots ($|Q|$ = 100) & & 0.696 s & 2.857 s & 1.642 s \\
        \quad \ldots ($|Q|$ = 200) & & 1.281 s & 5.699 s & 2.753 s \\
        \quad \ldots ($|Q|$ = 500) & & 2.763 s & 12.46 s & 4.626 s \\
        \quad \ldots ($|Q|$ = 1000) & & 4.808 s & 25.47 s & 7.957 s \\
        \midrule
        \textbf{Retrieving a text of length $d$} & &  & & & $O((a + b + d) H_0)$ \\
        \quad \ldots ($d$ = 10) & & 0.426 s & 0.895 s & 1.101 s  \\
        \quad \ldots ($d$ = 50) & & 0.634 s & 1.549 s & 1.302 s \\
        \quad \ldots ($d$ = 100) & & 0.734 s & 1.991 s & 1.326 s \\
        \quad \ldots ($d$ = 500) & & 0.874 s & 2.363 s & 1.609 s \\
        \quad \ldots ($d$ = 1000) & & 0.94 s & 2.385 s & 1.705 s \\
        \quad \ldots ($d$ = 2000) & & 1.213 s & 3.464 s & 2.849 s \\
        \quad \ldots ($d$ = 3000) & & 1.858 s & 4.456 s & 3.330 s \\
        \bottomrule
    \end{tabular}
    }%
    \caption{
    Inference time latency of \methodname{}. Average latency of each query is reported. 
    Notations: $n=$ number of bytes in the text corpus, $S=$ number of shards for the index, $|Q|=$ length of query in bytes, $d=$ length of text (in bytes) to reconstruct from the index, $a=$ sampling rate of SA, $b=$ sampling rate of ISA, $H_0=$ zeroth-order entropy of the corpus.
    }
    \label{tab:benchmarking}
\end{table*}

We benchmark the query latency of \methodname{}, and report results in \autoref{tab:benchmarking}.
For \textit{counting}, we experiment with query lengths $|Q| \in \{ 1, 2, 5, 10, 50, 100, 500, 1000 \}$, and for each length, randomly sample the query strings from the corpus.
For \textit{retrieving documents}, we use queries of length 10 sampled from the corpus, and reconstruct text surrounding the query with total length of $d \in \{10, 50, 100, 500, 1000\}$.
For each corpus and parameter setting, we repeat with 100 random queries and report the average latency.
For benchmarking, we store the indexes on \texttt{pd-balanced} SSD disks on GCP which has a max IOPS of 80,000 and max throughput of 1200 MB/s, and we experiment on an \texttt{n2-highcpu-64} node where the max disk I/O performance can be achieved.

\section{Query Latency Comparison with Prior Works}
\autoref{tab:infini-gram-latency} reports query latency on Pile-train corpus using our method compared to prior work by \citet{Liu2024InfiniGram}.

\begin{table*}[h]
\centering
\begin{tabular}{lccc}
\toprule
 & $|Q|=10$ bytes & $|Q|=20$ bytes & $|Q|=100$ bytes \\
\midrule
infini-gram \citep{Liu2024InfiniGram}   & 13 ms  & 14 ms  & 13 ms  \\
\textbf{\methodname{}}  & 106 ms & 182 ms & 696 ms \\
\bottomrule
\end{tabular}
\caption{Text retrieving latency comparison between infini-gram and \methodname{} on Pile-train corpus.}
\label{tab:infini-gram-latency}
\end{table*}

\section{\Methodname{} Web Interface}
\label{app:interfce}
\begin{figure*}
    \centering
    \includegraphics[width=\linewidth]{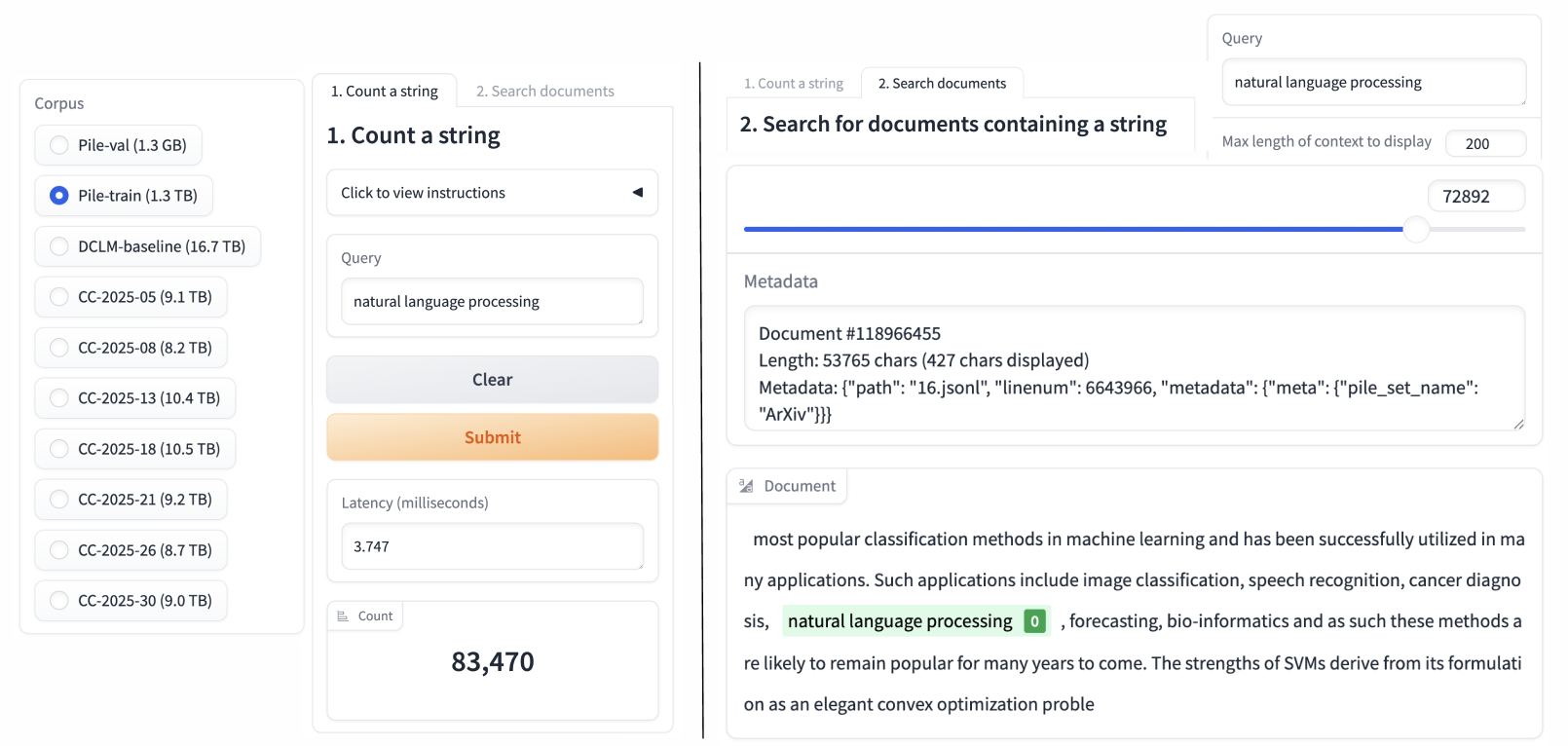}
    \caption{The web interface of \methodname{}. \textbf{Left:} counting a string. \textbf{Right:} retrieving documents.}
    \label{fig:web}
\end{figure*}

We host web interface for easy access to counting (\autoref{fig:web} left) and document retrieval (\autoref{fig:web} right) with the indexes we have built.

\section{Details for Benchmarks in Contamination Analysis}
\label{app:benchmark-citation}


\autoref{tab:benchmark-citation} shows the benchmarks we analyze under each category with citation and source. 
\begin{table*}[h!]
    \centering
    \resizebox{\textwidth}{!}{
    \begin{tabular}{lrl}
    \toprule
    \textbf{Benchmark} & \textbf{Citation} & \textbf{Source} \\
    \midrule
    
    \multicolumn{3}{c}{\textbf{Knowledge and Reasoning}} \\
    \midrule
    MMLU   & \citealp{mmlu} & \href{https://huggingface.co/datasets/cais/mmlu}{https://huggingface.co/datasets/cais/mmlu} \\
    MMLU-Pro   & \citealp{mmlu-pro} & \href{https://huggingface.co/datasets/TIGER-Lab/MMLU-Pro}{https://huggingface.co/datasets/TIGER-Lab/MMLU-Pro} \\
    BigBenchHard   & \citealp{bbh} & \href{https://github.com/suzgunmirac/BIG-Bench-Hard/tree/main/bbh}{https://github.com/suzgunmirac/BIG-Bench-Hard/tree/main/bbh} \\
    AGIEval   & \citealp{agieval} & \href{https://github.com/ruixiangcui/AGIEval/tree/main/data/v1_1}{https://github.com/ruixiangcui/AGIEval/tree/main/data/v1\_1} \\
    GPQA   & \citealp{gpqa} & \href{https://huggingface.co/datasets/Idavidrein/gpqa}{https://huggingface.co/datasets/Idavidrein/gpqa} \\
    HLE   & \citealp{phan2025humanitysexam} & \href{https://huggingface.co/datasets/cais/hle}{https://huggingface.co/datasets/cais/hle} \\
    \midrule
    
    \multicolumn{3}{c}{\textbf{Math}} \\
    \midrule
    AIME-2024   & \citealp{AoPS2024AIMEI} & \href{https://huggingface.co/datasets/Maxwell-Jia/AIME_2024}{https://huggingface.co/datasets/Maxwell-Jia/AIME\_2024} \\
    GSM8K   & \citealp{cobbe2021gsm8k} & \href{https://huggingface.co/datasets/openai/gsm8k}{https://huggingface.co/datasets/openai/gsm8k} \\
    MATH-500   & \citealp{math-500} & \href{https://huggingface.co/datasets/HuggingFaceH4/MATH-500}{https://huggingface.co/datasets/HuggingFaceH4/MATH-500} \\
    MGSM   & \citealp{mgsm} & \href{https://huggingface.co/datasets/juletxara/mgsm}{https://huggingface.co/datasets/juletxara/mgsm} \\
    \midrule
    
    \multicolumn{3}{c}{\textbf{Code}} \\
    \midrule
    HumanEval   & \citealp{humaneval} & \href{https://huggingface.co/datasets/openai/openai_humaneval}{https://huggingface.co/datasets/openai/openai\_humaneval} \\
    HumanEval+   & \citealp{evalplus} & \href{https://huggingface.co/datasets/evalplus/humanevalplus}{https://huggingface.co/datasets/evalplus/humanevalplus} \\
    LiveCodeBench & \citealp{livecodebench} & \href{https://huggingface.co/datasets/livecodebench/code_generation}{https://huggingface.co/datasets/livecodebench/code\_generation} \\
    \quad (code generation)   &  \\
    SWE-bench   & \citealp{jimenez2024swebenchlanguagemodelsresolve} & \href{https://huggingface.co/datasets/princeton-nlp/SWE-bench_Verified}{https://huggingface.co/datasets/princeton-nlp/SWE-bench\_Verified} \\
    MBPP   & \citealp{mbpp} & \href{https://huggingface.co/datasets/google-research-datasets/mbpp}{https://huggingface.co/datasets/google-research-datasets/mbpp} \\
    \midrule
    
    \multicolumn{3}{c}{\textbf{Commonsense Understanding}} \\
    \midrule
    ARC-Challenge   & \citealp{allenai:arc} & \href{https://huggingface.co/datasets/allenai/ai2_arc}{https://huggingface.co/datasets/allenai/ai2\_arc} \\
    ARC-Easy   & \citealp{allenai:arc} & \href{https://huggingface.co/datasets/allenai/ai2_arc}{https://huggingface.co/datasets/allenai/ai2\_arc} \\
    CSQA   & \citealp{talmor-etal-2019-commonsenseqa} & \href{https://huggingface.co/datasets/tau/commonsense_qa}{https://huggingface.co/datasets/tau/commonsense\_qa} \\
    HellaSwag   & \citealp{zellers2019hellaswag} & \href{https://huggingface.co/datasets/Rowan/hellaswag}{https://huggingface.co/datasets/Rowan/hellaswag} \\
    Openbook QA   & \citealp{OpenBookQA2018} & \href{https://huggingface.co/datasets/allenai/openbookqa}{https://huggingface.co/datasets/allenai/openbookqa} \\
    Social IQa   & \citealp{sap2019socialiqacommonsensereasoningsocial} & \href{https://huggingface.co/datasets/allenai/social_i_qa}{https://huggingface.co/datasets/allenai/social\_i\_qa} \\
    WinoGrande   & \citealp{sakaguchi2019winograndeadversarialwinogradschema} & \href{https://huggingface.co/datasets/allenai/winogrande}{https://huggingface.co/datasets/allenai/winogrande} \\
    \midrule

    \multicolumn{3}{c}{\textbf{Reading Comprehension}} \\
    \midrule
    CoQA   & \citealp{reddy-etal-2019-coqa} & \href{https://huggingface.co/datasets/stanfordnlp/coqa}{https://huggingface.co/datasets/stanfordnlp/coqa} \\
    SQuAD   & \citealp{rajpurkar-etal-2016-squad} & \href{https://huggingface.co/datasets/rajpurkar/squad}{https://huggingface.co/datasets/rajpurkar/squad} \\

    \bottomrule
    \end{tabular}
    }
    \caption{Citation and source of each benchmark analyzed in this paper.
    }
    \label{tab:benchmark-citation}
\end{table*}


\section{Detailed Benchmark Contamination Result}
\label{app:more-contamination-result}

\autoref{tab:full_benchmark} reports the number of suspicious and dirty entries of each benchmark.

\clearpage
\begin{sidewaystable*}
    \centering
    \resizebox{\paperwidth}{!}{%
    \begin{tabular}{l|r|rr|rr|rr|rr|rr|rr|rr|rr}
        \toprule
        \textbf{Benchmark} & \textbf{Test Set} & \multicolumn{2}{c}{\textbf{Pile-train}} & \multicolumn{2}{c}{\textbf{DCLM-baseline}} & \multicolumn{2}{c}{\textbf{CC-2025-05}} & \multicolumn{2}{c}{\textbf{CC-202508}} & \multicolumn{2}{c}{\textbf{CC-202513}} & \multicolumn{2}{c}{\textbf{CC-202518}} & \multicolumn{2}{c}{\textbf{CC-202521}} & \multicolumn{2}{c}{\textbf{CC-202526}} \\
        \cmidrule(lr){3-4} \cmidrule(lr){5-6} \cmidrule(lr){7-8} \cmidrule(lr){9-10} \cmidrule(lr){11-12} \cmidrule(lr){13-14} \cmidrule(lr){15-16}
         & \textbf{Size} & \textbf{Sus} & \textbf{Dirty} & \textbf{Sus} & \textbf{Dirty} & \textbf{Sus} & \textbf{Dirty} & \textbf{Sus} & \textbf{Dirty} & \textbf{Sus} & \textbf{Dirty} & \textbf{Sus} & \textbf{Dirty} & \textbf{Sus} & \textbf{Dirty} & \textbf{Sus} & \textbf{Dirty}\\
        \midrule
        \multicolumn{18}{c}{\textbf{Knowledge and Reasoning}} \\
        \midrule
        \textbf{MMLU} & 1000  & 57 & 133 & 142 & 277 & 122 & 135 & 139 & 90 & 121 & 122 & 115 & 120 & 117 & 125 & 92 & 124 \\
        \textbf{MMLU-Pro} & 1000  & 35 & 59 & 77 & 162 & 31 & 81 & 55 & 54 & 60 & 52 & 63 & 58 & 74 & 60 & 69 & 40 \\
        \textbf{BigBenchHard} & 1000  & 7 & 0 & 54 & 1 & 357 & 14 & 296 & 14 & 306 & 32 & 302 & 23 & 396 & 18 & 338 & 17 \\
        \textbf{AGIEval} & 1000  & 45 & 8 & 155 & 31 & 107 & 27 & 92 & 36 & 115 & 30 & 109 & 70 & 95 & 94 & 104 & 46 \\
        \textbf{GPQA}  & 448 & 1 & 0 & 5 & 0 & 6 & 4 & 10 & 9 & 5 & 6 & 6 & 3 & 5 & 4 & 15 & 12 \\
        \textbf{HLE}  & 881 & 3 & 0 & 6 & 0 & 8 & 0 & 3 & 0 & 4 & 1 & 5 & 0 & 3 & 0 & 2 & 0 \\
        
        \midrule
        \multicolumn{18}{c}{\textbf{Math}} \\
        \midrule
        \textbf{AIME-2024}  & 30  & 2 & 0 & 4 & 0 & 13 & 3 & 3 & 1 & 10 & 2 & 13 & 12 & 13 & 12 & 12 & 4 \\
        \textbf{GSM8K} & 1000  & 1 & 0 & 5 & 4 & 33 & 76 & 8 & 8 & 37 & 69 & 10 & 7 & 210 & 742 & 20 & 73 \\
        \textbf{MATH-500}  & 500  & 9 & 3 & 48 & 16 & 26 & 3 & 46 & 39 & 25 & 4 & 27 & 4 & 26 & 4 & 41 & 41 \\
        \textbf{MGSM} & 250  & 0 & 0 & 2 & 0 & 9 & 14 & 3 & 4 & 37 & 89 & 5 & 2 & 54 & 182 & 6 & 15 \\
        \midrule
        \multicolumn{18}{c}{\textbf{Code}} \\
        \midrule
        \textbf{HumanEval} & 164  & 1 & 0 & 1 & 0 & 14 & 0 & 60 & 1 & 53 & 1 & 55 & 1 & 65 & 0 & 77 & 0 \\
        \textbf{HumanEval+}  & 164 & 1 & 0 & 1 & 0 & 14 & 0 & 60 & 1 & 53 & 1 & 55 & 1 & 65 & 0 & 77 & 0 \\
        \textbf{LiveCodeBench} & 880  & 2 & 0 & 2 & 0 & 188 & 0 & 241 & 0 & 221 & 0 & 274 & 0 & 167 & 0 & 168 & 0 \\
        \textbf{SWE-bench}  & 500  & 5 & 0 & 26 & 0 & 29 & 1 & 38 & 1 & 32 & 0 & 34 & 2 & 33 & 0 & 28 & 0 \\
        \textbf{MBPP} & 500 & 16 & 0 & 64 & 2 & 87 & 5 & 58 & 7 & 89 & 6 & 75 & 9 & 76 & 5 & 84 & 7 \\
        \midrule
        \multicolumn{18}{c}{\textbf{Commonsense Understanding}} \\
        \midrule
        \textbf{ARC-Challenge} & 1000  & 6 & 14 & 4 & 341 & 18 & 119 & 29 & 40 & 27 & 31 & 26 & 38 & 28 & 42 & 29 & 48 \\
        \textbf{ARC-Easy}  & 1000  & 15 & 15 & 5 & 317 & 31 & 54 & 30 & 95 & 31 & 55 & 37 & 55 & 28 & 61 & 31 & 62 \\
        \textbf{CSQA}  & 1000 & 0 & 1 & 1 & 10 & 1 & 1 & 0 & 1 & 0 & 2 & 0 & 1 & 0 & 0 & 0 & 1 \\
        \textbf{HellaSwag} & 1000 & 27 & 0 & 341 & 0 & 36 & 0 & 29 & 0 & 34 & 0 & 29 & 0 & 30 & 0 & 28 & 1 \\
        \textbf{OpenbookQA} & 500 & 2 & 54 & 3 & 78 & 5 & 73 & 2 & 151 & 2 & 66 & 3 & 67 & 3 & 66 & 2 & 61 \\
        \textbf{Social IQa} & 1000 & 0 & 0 & 2 & 5 & 2 & 2 & 1 & 44 & 2 & 2 & 2 & 3 & 3 & 2 & 0 & 1 \\
        \textbf{WinoGrande}  & 1000 & 2 & 0 & 0 & 0 & 0 & 0 & 0 & 0 & 0 & 0 & 0 & 0 & 0 & 0 & 0 & 0 \\
        \midrule
        \multicolumn{18}{c}{\textbf{Reading Comprehension}} \\
        \midrule
        \textbf{CoQA}  & 500 & 189 & 40 & 188 & 92 & 121 & 37 & 113 & 44 & 123 & 43 & 134 & 36 & 123 & 38 & 125 & 44 \\
        \textbf{SQuAD}  & 1000 & 16 & 28 & 9 & 401 & 11 & 27 & 12 & 330 & 15 & 101 & 13 & 15 & 12 & 20 & 11 & 85 \\
        \bottomrule
    \end{tabular}
    }%
    \caption{Detailed benchmark contamination results. For test sets with more than 1000 entries, we randomly downsample to 1000.}
    \label{tab:full_benchmark}
\end{sidewaystable*}
\clearpage

\section{Contamination Examples}
\label{app:examples}

\begin{figure*}[h!]
    \centering
    \begin{subfigure}{\linewidth}
        \includegraphics[width=1\linewidth]{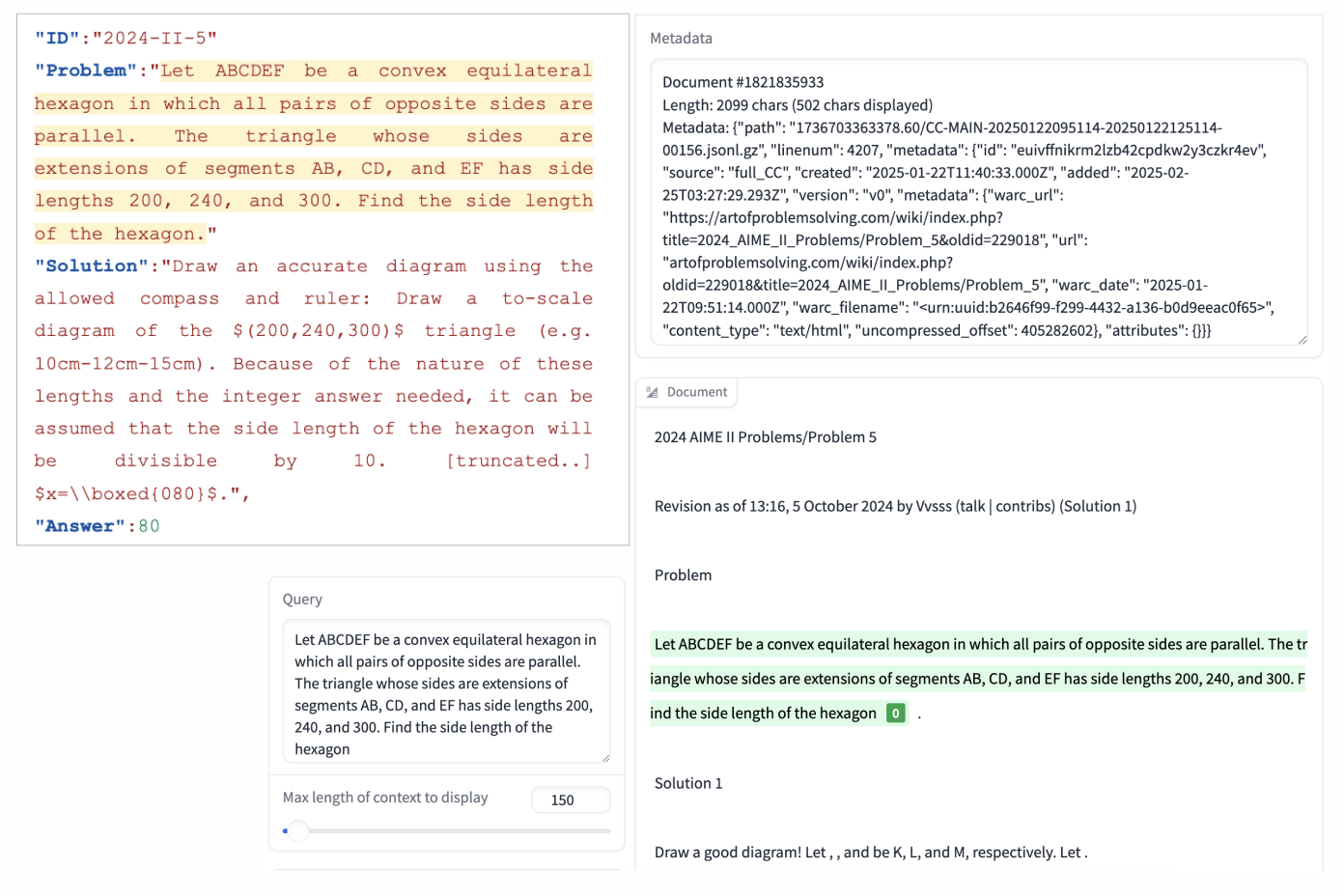}
        \caption{\textbf{Left upper:} An entry in the AIME-2024 benchmark. \textbf{Right:} A document contaminating this entry, retrieved from CC-2025-05 by \methodname{}. This example belongs to Category 1, where the correct answer can be found in the document.}
    \end{subfigure}
    \begin{subfigure}{\linewidth}
        \centering
        \includegraphics[width=0.9\linewidth]{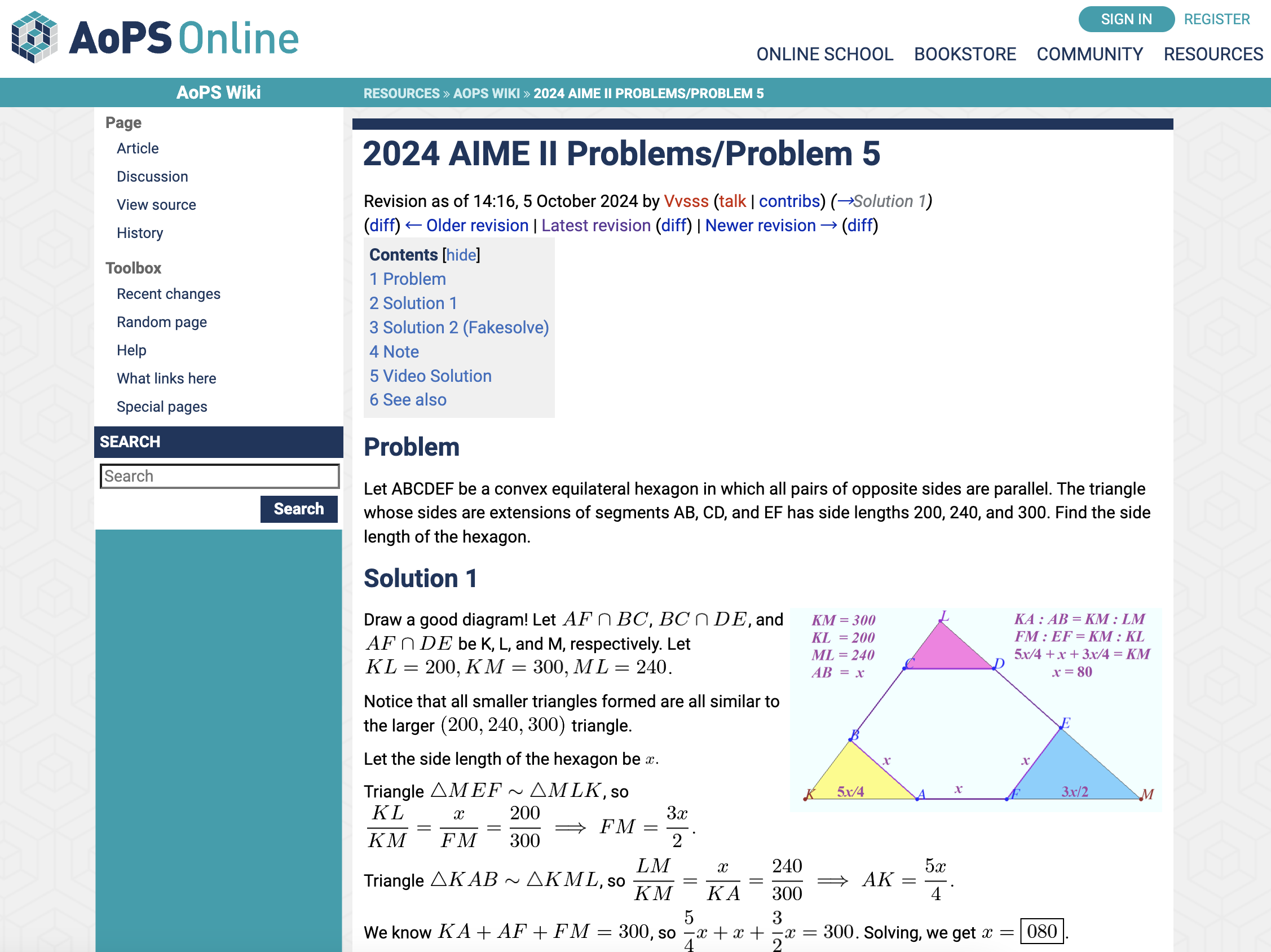}
        \caption{The original webpage responsible for the contamination.}
    \end{subfigure}
    \caption{AIME dirty entry example in CC-2025-05. The contamination source is the official AOPS website, where AIME exams are published.}
    \label{fig:example_aime}
\end{figure*}

\begin{figure*}[h!]
    \centering
    \begin{subfigure}{\linewidth}
        \includegraphics[width=1\linewidth]{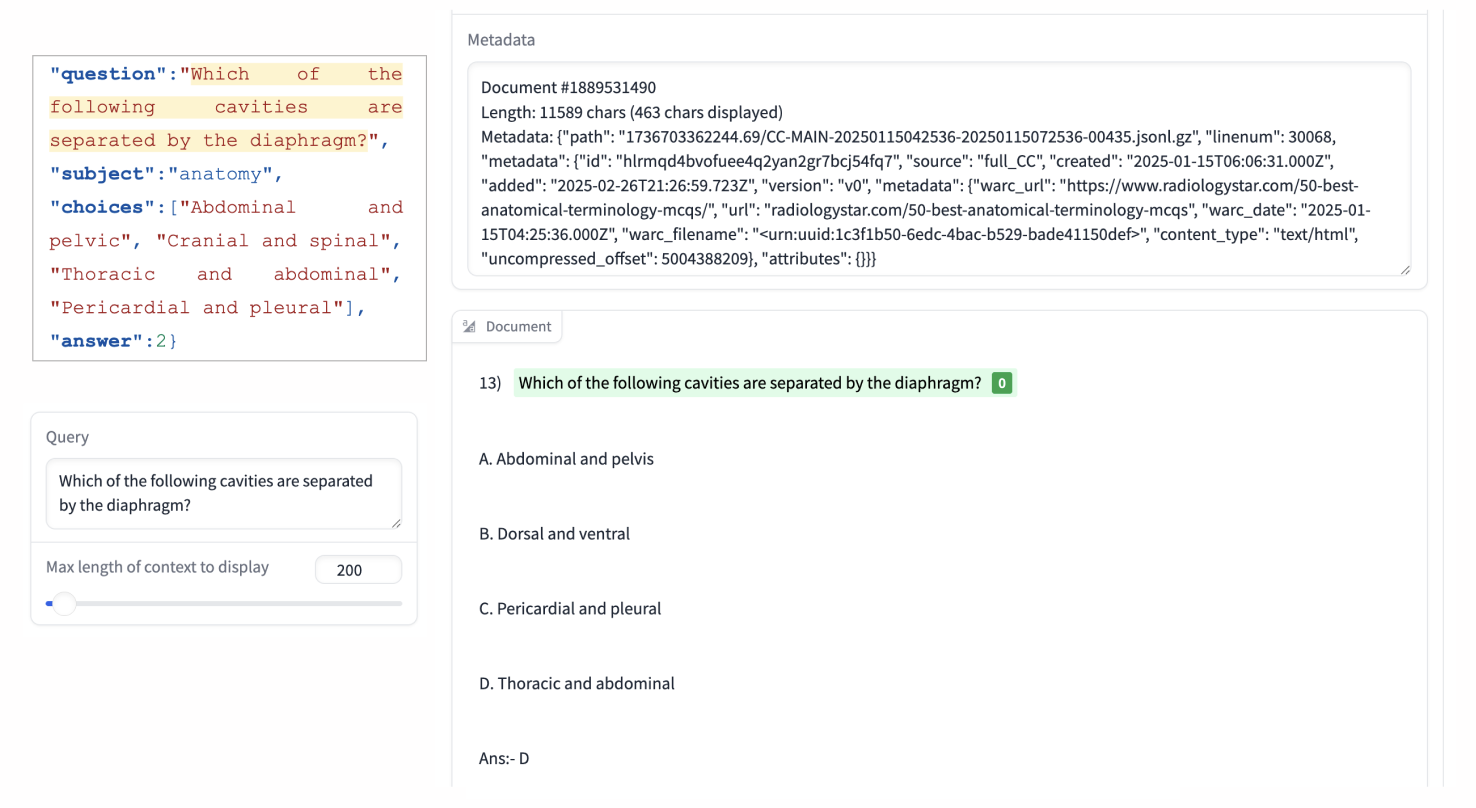}
        \caption{\textbf{Left upper:} An entry in the MMLU benchmark. \textbf{Right:} A document contaminating this entry, retrieved from CC-2025-05 by \methodname{}. This example belongs to Category 1, where the correct answer presents (though choices are not in the exact same order).}
    \end{subfigure}
    \begin{subfigure}{\linewidth}
        \centering
        \includegraphics[width=\linewidth]{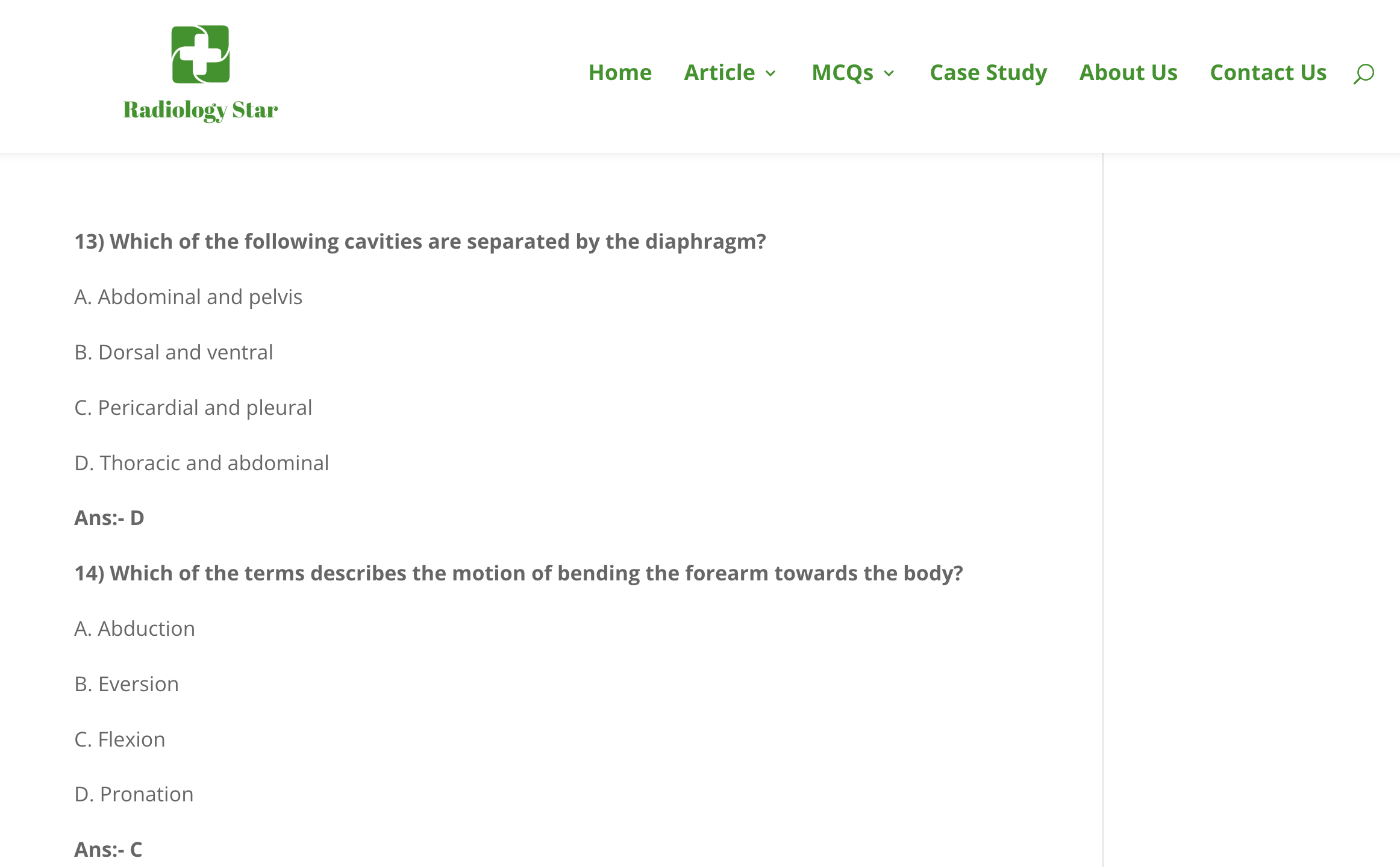}
        \caption{The original webpage responsible for the contamination.}
    \end{subfigure}
    \caption{MMLU dirty entry example in CC-2025-05. The contamination source is a website containing multiple-choice question in related fields.}
    \label{fig:example_mmlu}
\end{figure*}

\begin{figure*}[h!]
    \centering
    \begin{subfigure}{\linewidth}
        \centering
        \includegraphics[width=0.9\linewidth]{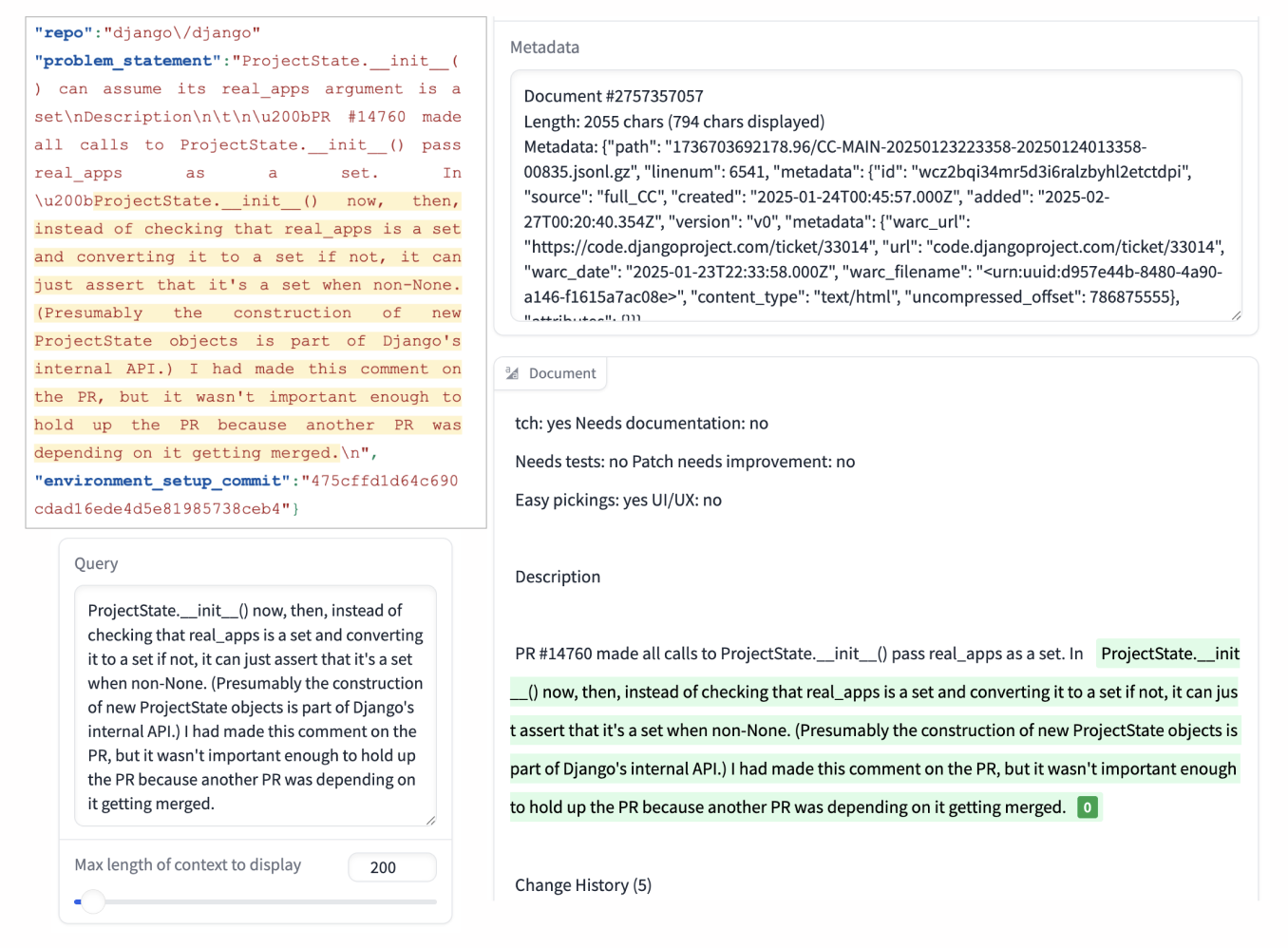}
        \caption{\textbf{Left upper:} An entry in the SWE-bench benchmark. \textbf{Right:} A document contaminating this entry, retrieved from CC-2025-05 by \methodname{}. This example is Category 3, where only question appears but not the answer.}
    \end{subfigure}
    \begin{subfigure}{\linewidth}
        \centering
        \includegraphics[width=0.8\linewidth]{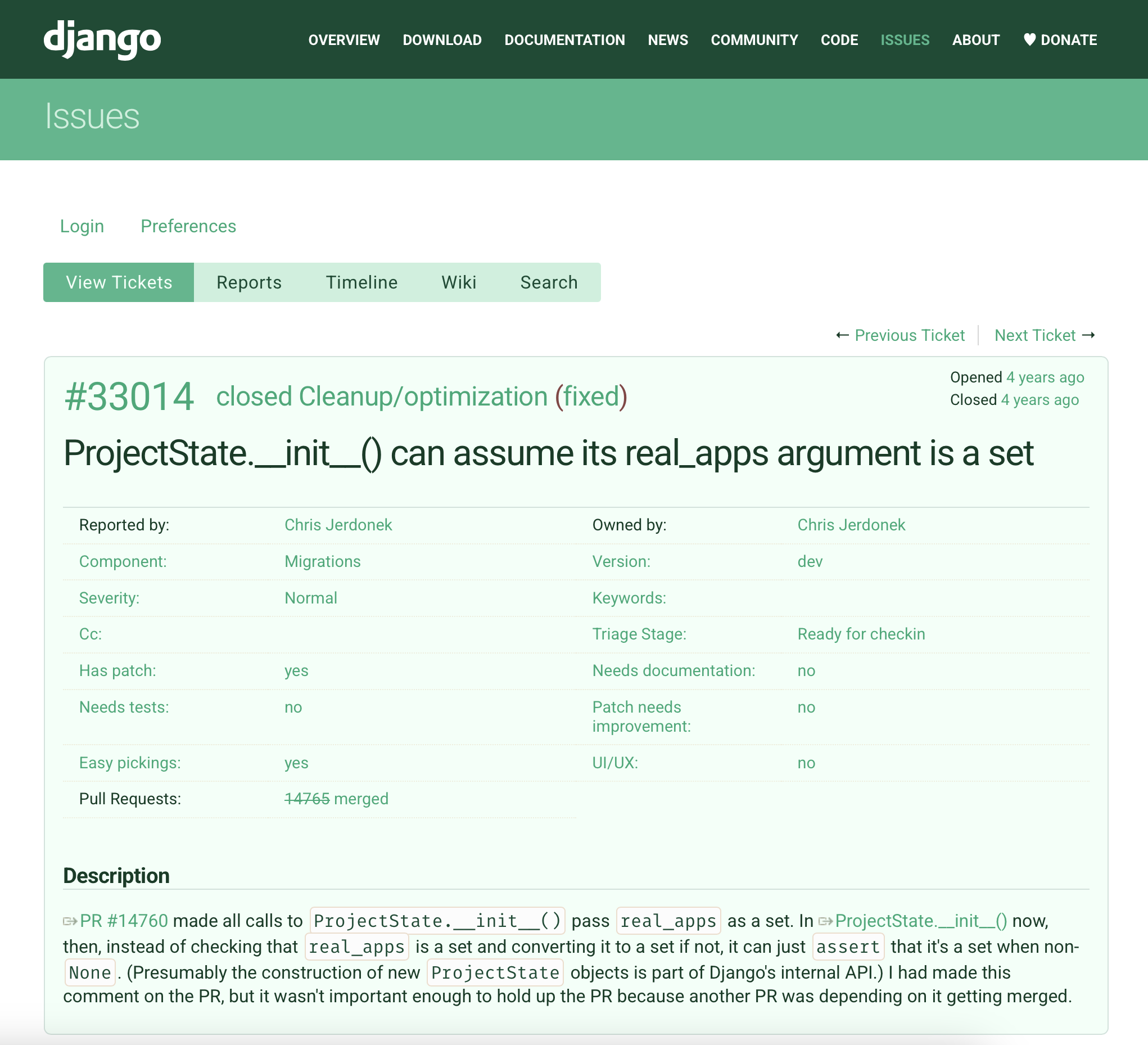}
        \caption{The original webpage responsible for the contamination.}
    \end{subfigure}
    \caption{SWE-bench dirty entry example in CC-2025-05. The contamination source is a website recording pull requests for software developing.}
    \label{fig:example_swebench}
\end{figure*}

\begin{figure*}[h!]
    \centering
    \begin{subfigure}{\linewidth}
        \includegraphics[width=1\linewidth]{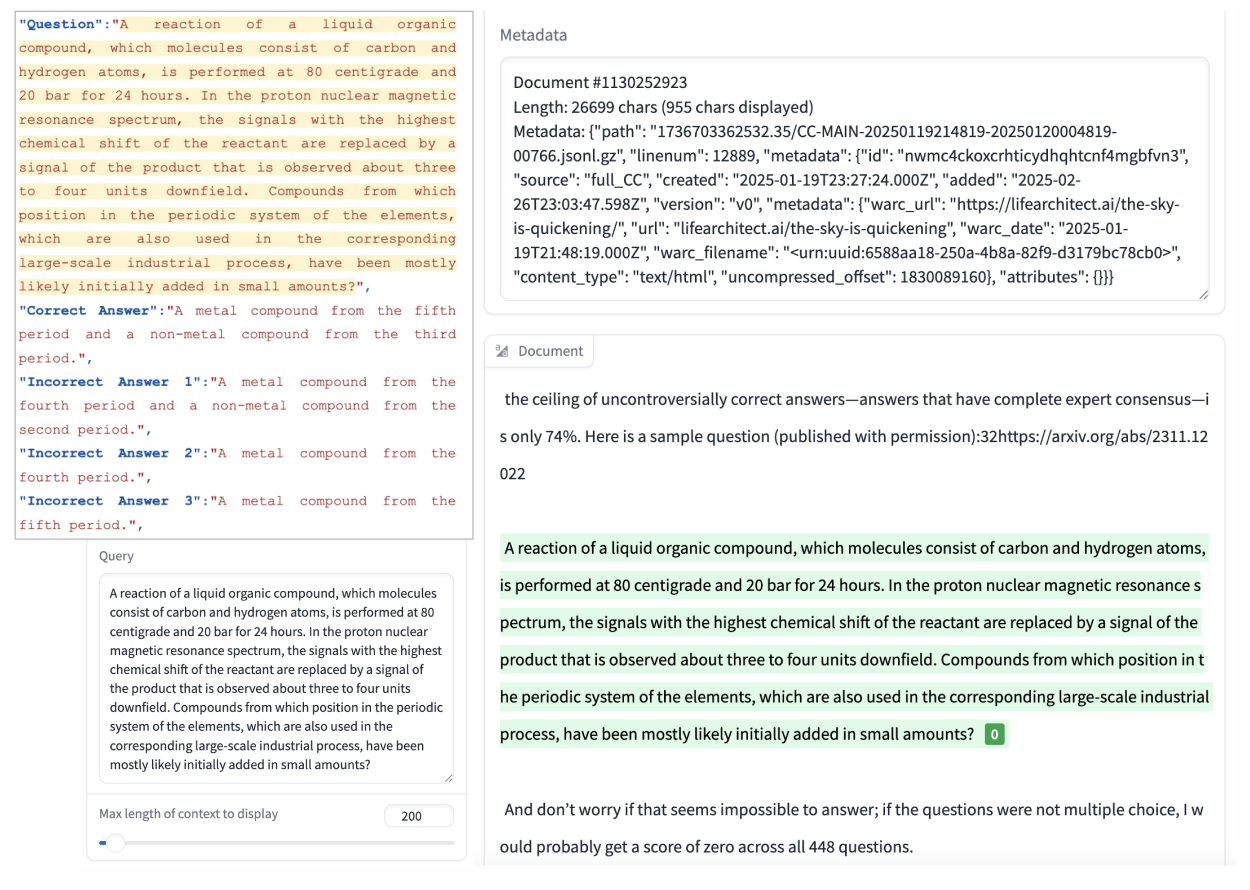}
        \caption{\textbf{Left bottom:} An entry in the GPQA benchmark. \textbf{Right:} A document contaminating this entry, retrieved from CC-2025-05 by \methodname{}. This example belongs to Category 3, where only the question appears, not the answers.}
    \end{subfigure}
    \begin{subfigure}{\linewidth}
        \centering
        \includegraphics[width=0.8\linewidth]{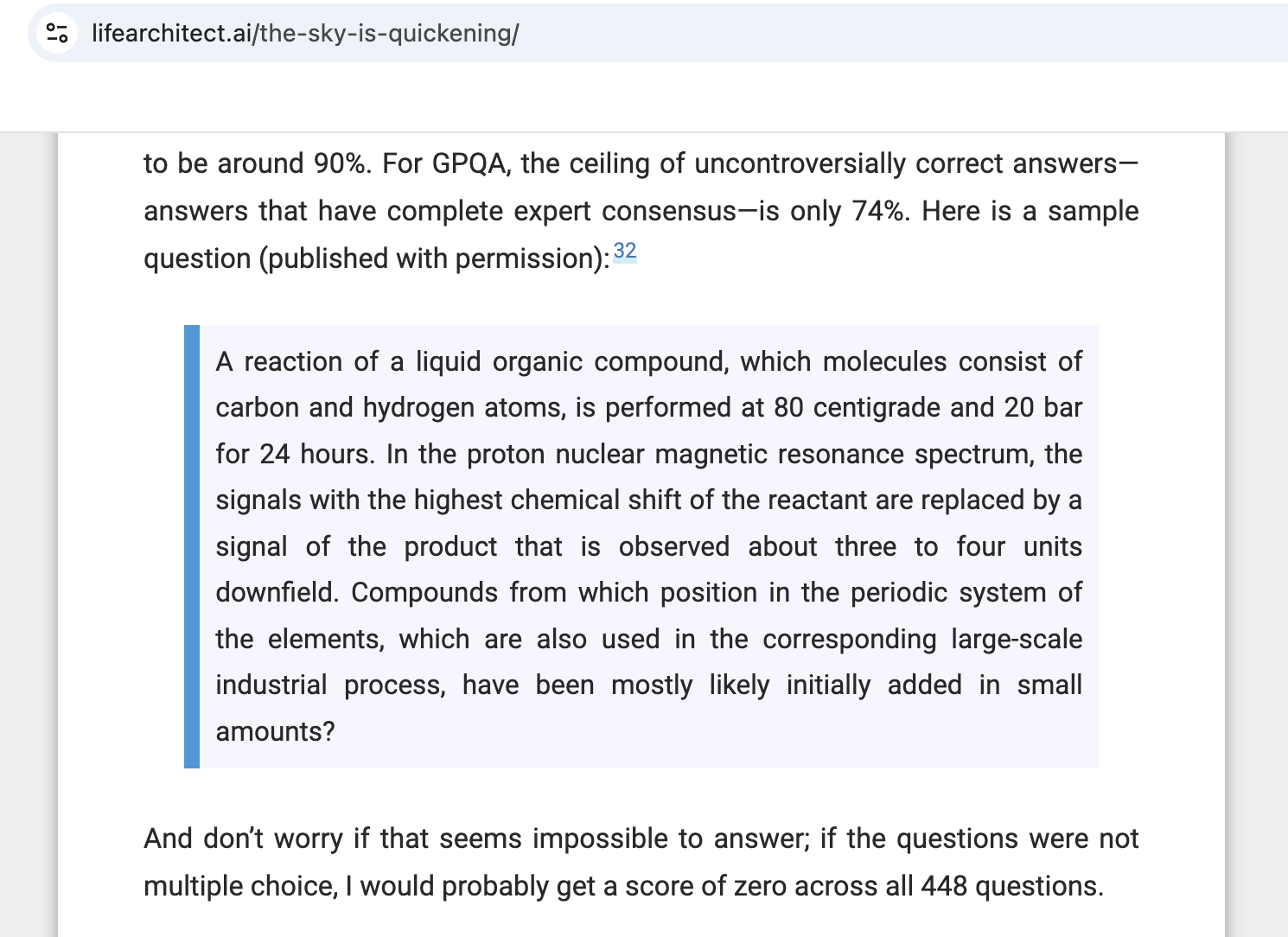}
        \caption{The original webpage responsible for the contamination.}
    \end{subfigure}
    \caption{GPQA dirty entry example in CC-2025-05. The contamination source is a blog post citing a test set example.}
    \label{fig:example_gpqa}
\end{figure*}

\begin{figure*}[h!]
    \centering
    \begin{subfigure}{\linewidth}
        \includegraphics[width=1\linewidth]{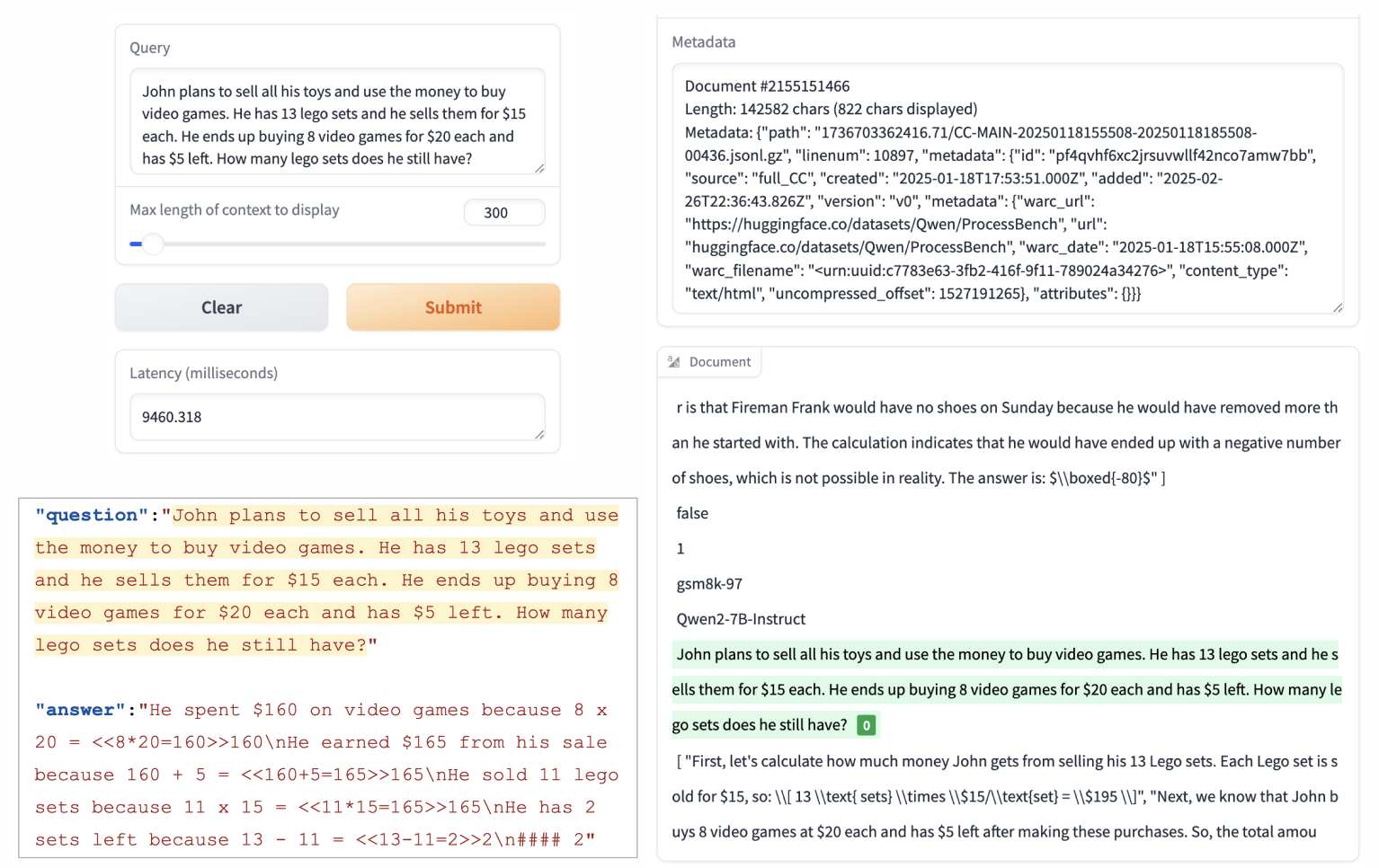}
        \caption{\textbf{Left bottom:} An entry in the GSM8K benchmark. \textbf{Right:} A document contaminating this entry, retrieved from CC-2025-05 by \methodname{}. This example is Category 3 because the correct answer is not present. }
    \end{subfigure}
    \begin{subfigure}{\linewidth}
        \includegraphics[width=1\linewidth]{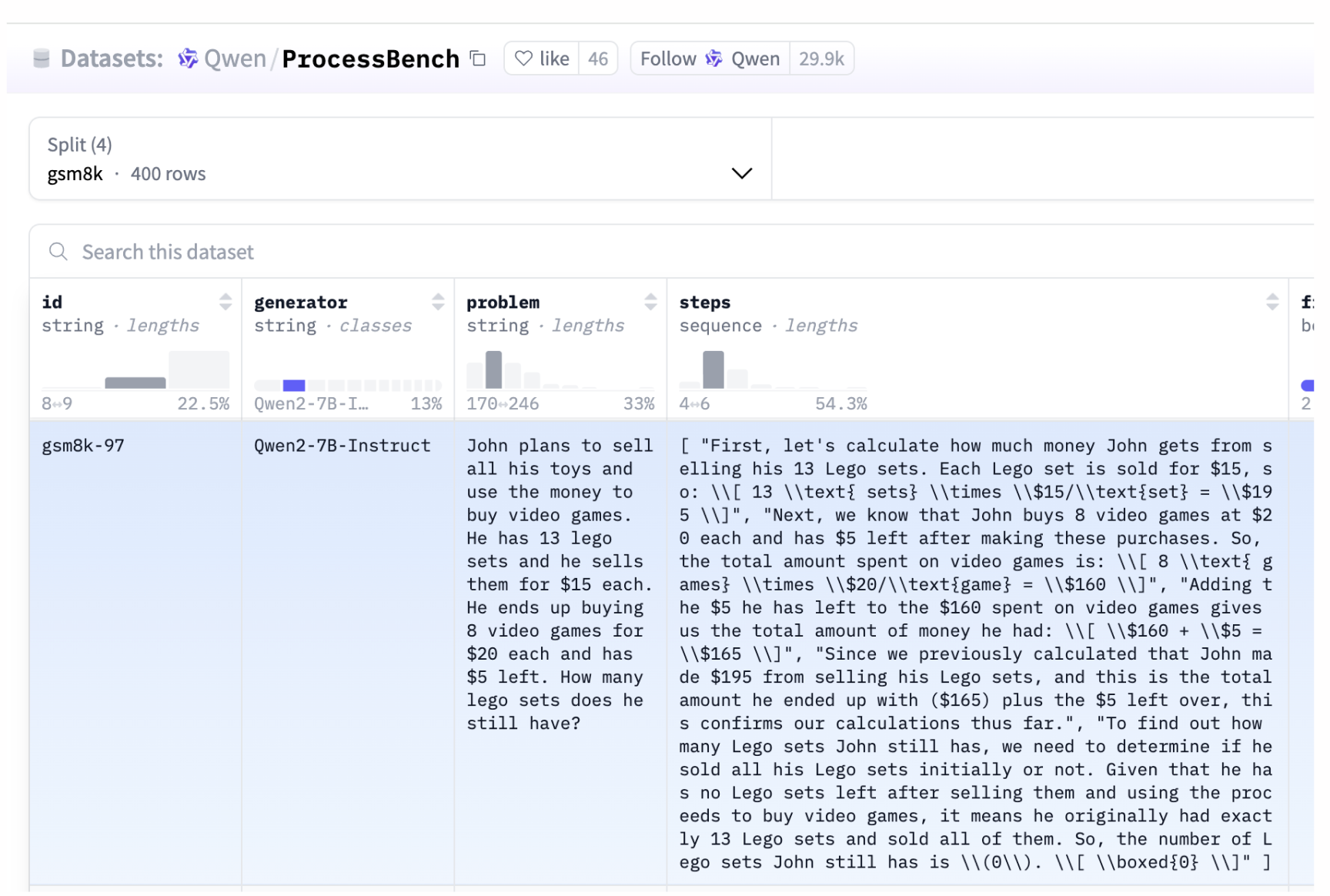}
        \caption{The original webpage responsible for the contamination.}
    \end{subfigure}
    \caption{GSM8K dirty entry example in CC-2025-05. The contamination source is a HuggingFace dataset sourcing from GSM8K examples, and is the major source for GSM8K dirty entries. This new dataset contains erroneous ``steps'' field and final answer to examine LLM's ability to identify errors, so the correct answers do not appear.}
    \label{fig:example_gsm8k}
\end{figure*}

\begin{figure*}[h!]
    \centering
    \begin{subfigure}{\linewidth}
        \includegraphics[width=1\linewidth]{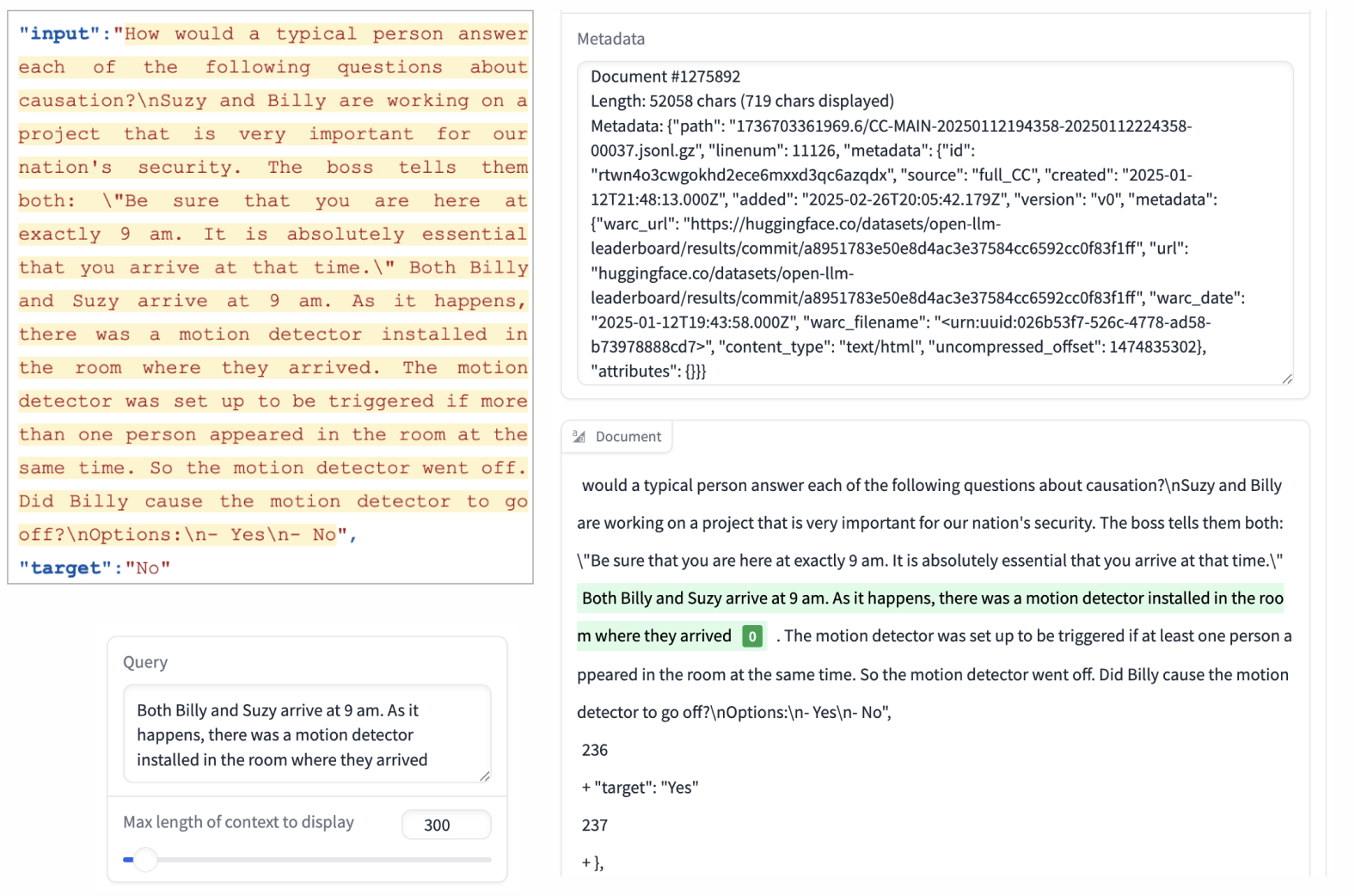}
        \caption{\textbf{Left bottom:} An entry in the BigBenchHard benchmark. \textbf{Right:} A document contaminating this entry, retrieved from CC-2025-05 by \methodname{}. This example is Category 3 because the correct answer is not present.}
    \end{subfigure}
    \begin{subfigure}{\linewidth}
        \includegraphics[width=1\linewidth]{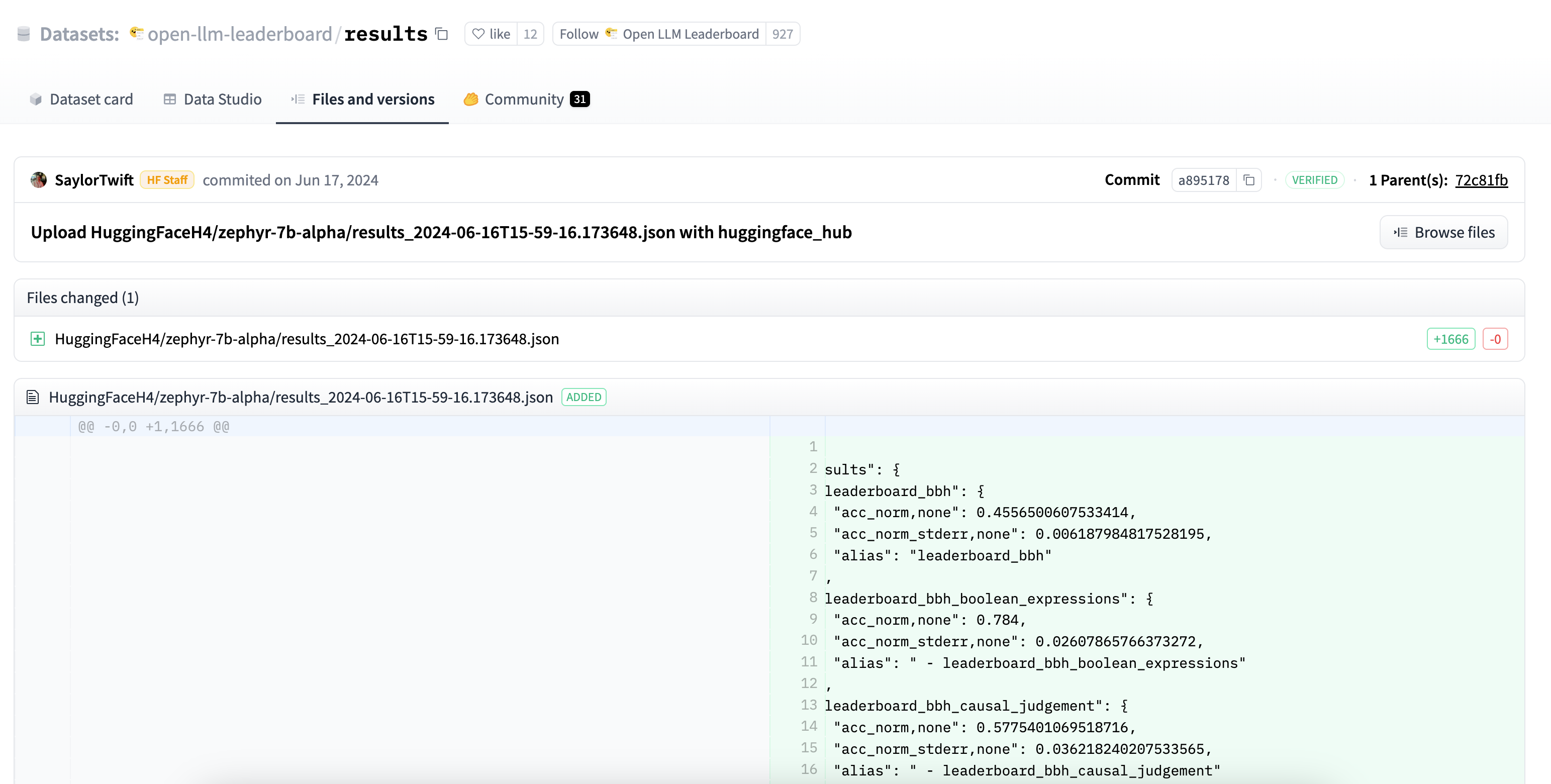}
        \caption{The original webpage responsible for the contamination.}
    \end{subfigure}
    \caption{BigBenchHard dirty entry example in CC-2025-05. The contamination source is a HuggingFace commit history that list it as few-shot example.}
    \label{fig:example_bbh}
\end{figure*}

\begin{figure*}[h!]
    \centering
    \begin{subfigure}{\linewidth}
        \includegraphics[width=1\linewidth]{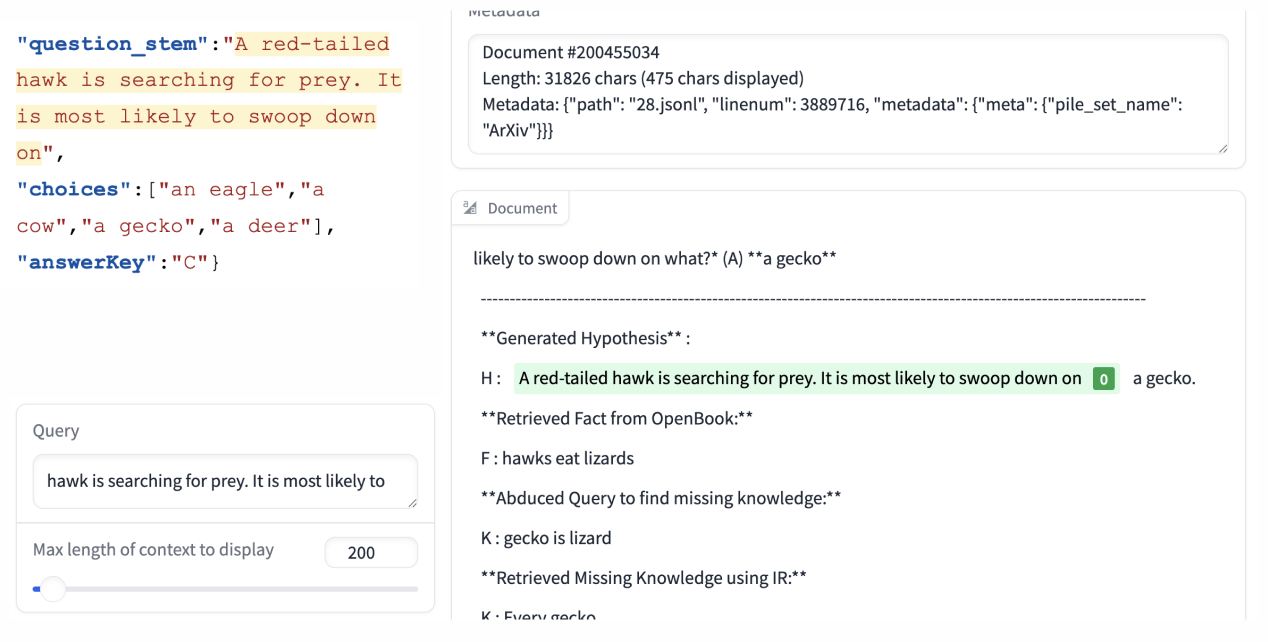}
        \caption{\textbf{Left bottom:} An entry in the OpenbookQA benchmark. \textbf{Right:} A document contaminating this entry, retrieved from Pile-train by \methodname{}. This example is Category 2 where the correct answer appears in natural language.}
    \end{subfigure}
    \begin{subfigure}{\linewidth}
        \includegraphics[width=1\linewidth]{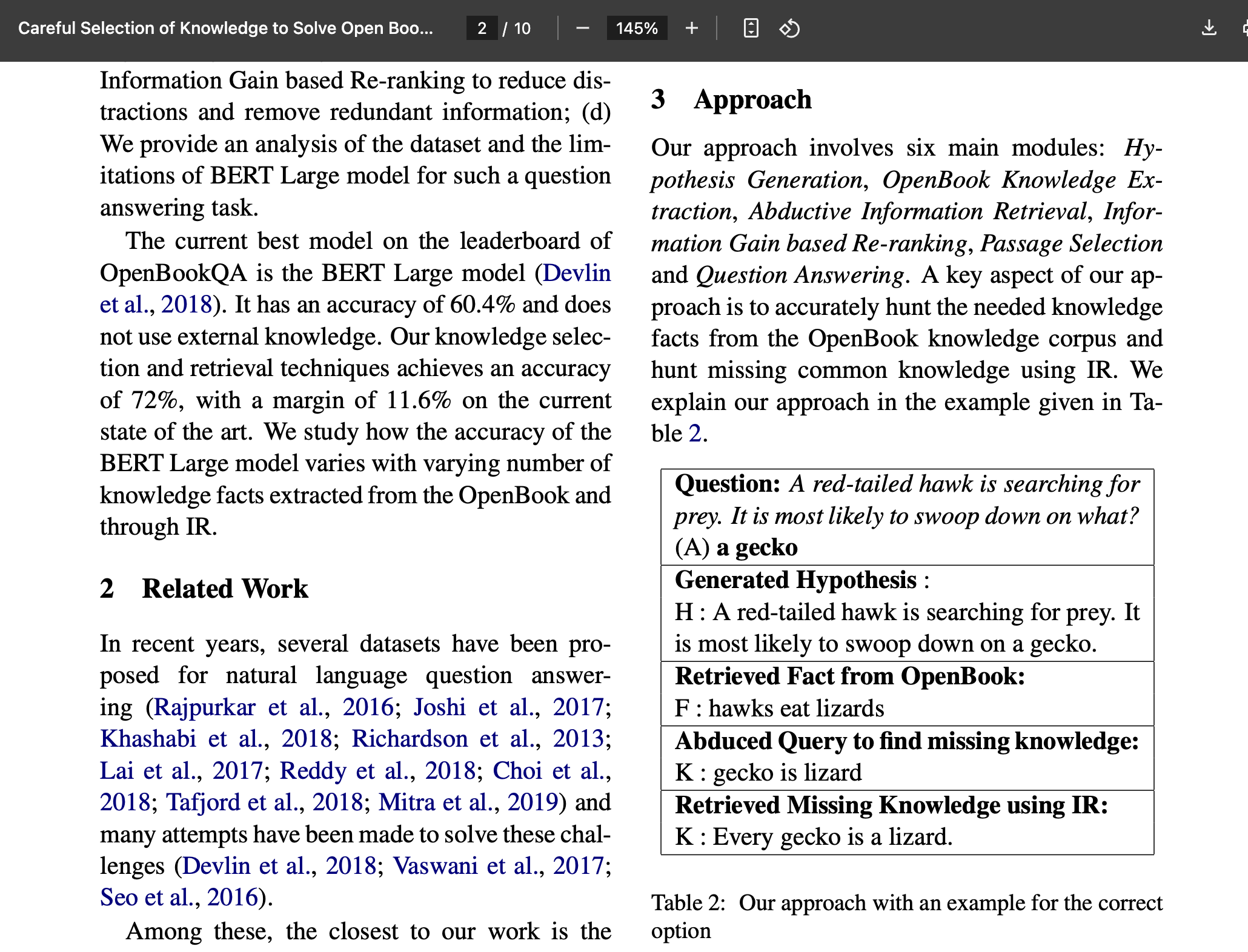}
        \caption{The original webpage responsible for the contamination. }
    \end{subfigure}
    \caption{OpenbookQA dirty entry example in Pile-train. The contamination source is a paper citing an example from OpenbookQA benchmark.}
    \label{fig:example_obqa}
\end{figure*}

\autoref{fig:example_aime} to \autoref{fig:example_obqa} shows example dirty entries in seven benchmarks and its contamination source retrieved from one of three corpus using \methodname{}. We also present the original webpage that is responsible for the contamination.



\section{Dirty Entry Categorization using LLM-as-a-judge}
\label{app:annotation}
We use gpt-4o-mini to categorize all dirty entries into one of four categories described in \S\ref{sec:contamination-analysis}. For each dirty entry, we extract the 50-character substring that has least occurrence in the corpus. We then retrieve text snippet with context length of 400 characters (850 characters in total). We prompt gpt-4o-mini by providing both the dirty entry and the text snippet and let the model decide which category the dirty entry belongs. We use the following prompt:

\textsf{You are an expert in evaluating benchmark contamination. Given an entry and detected overlap in the corpus, categorize how the entry is contaminated in corpus into one of four categories: \newline 1. You can find exact match of question stem and the correct answer (correct choice if multiple choices, or answer matching exactly the answer field) in corpus. \newline 2. You can find exact match of question stem, and the correct answer appears, though not in exact match. \newline 3. You can find exact match of question stem, but you cannot find the correct answer in any form. \newline 4. False positive \newline Output only the category number.} 

\section{Interface of the Benchmark Contamination Monitoring System}
\label{app:bulletin-interface}

\begin{figure*}[h!]
    \centering
    \includegraphics[width=1\linewidth]{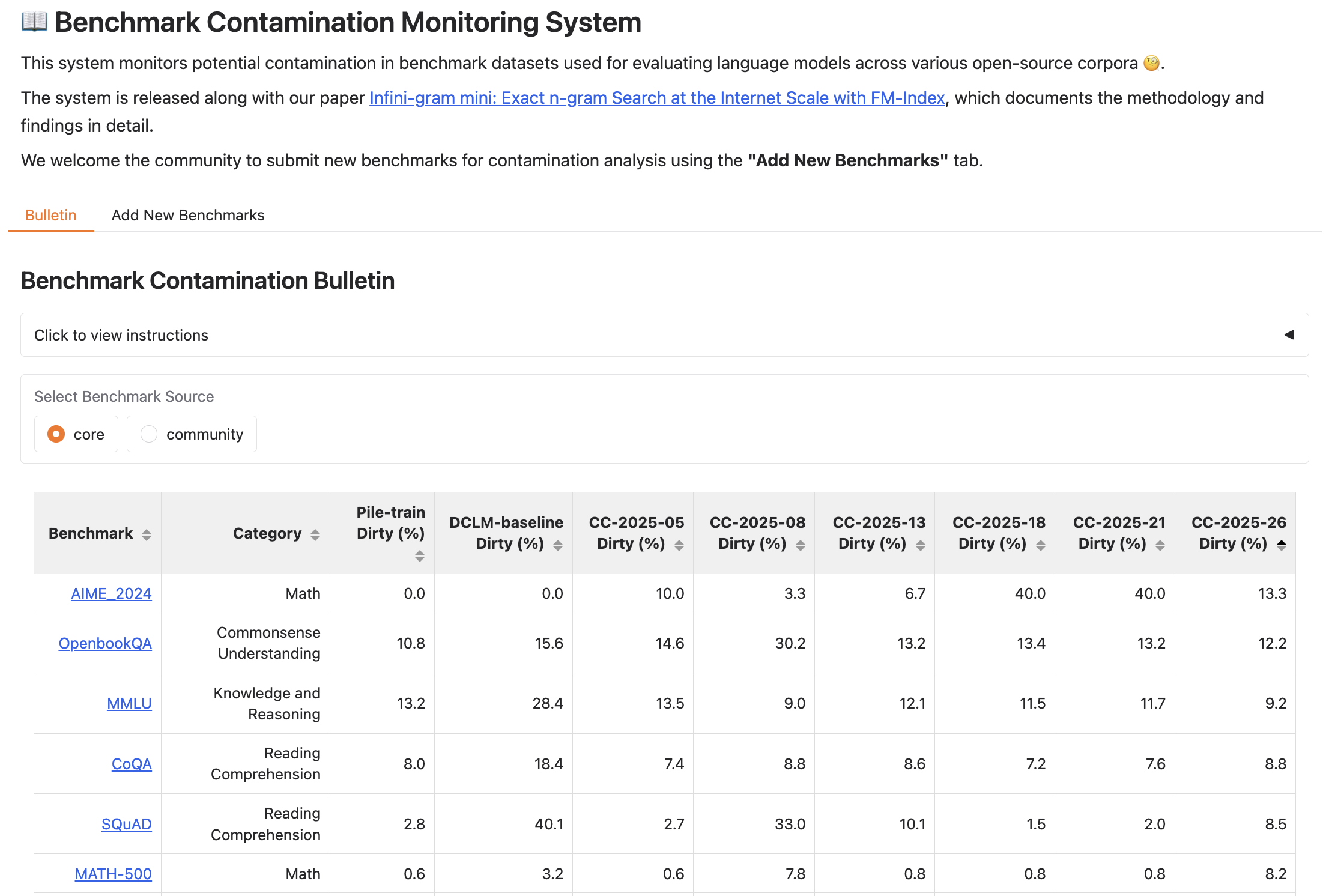} \\
    \vspace{16pt}
    \includegraphics[width=1\linewidth]{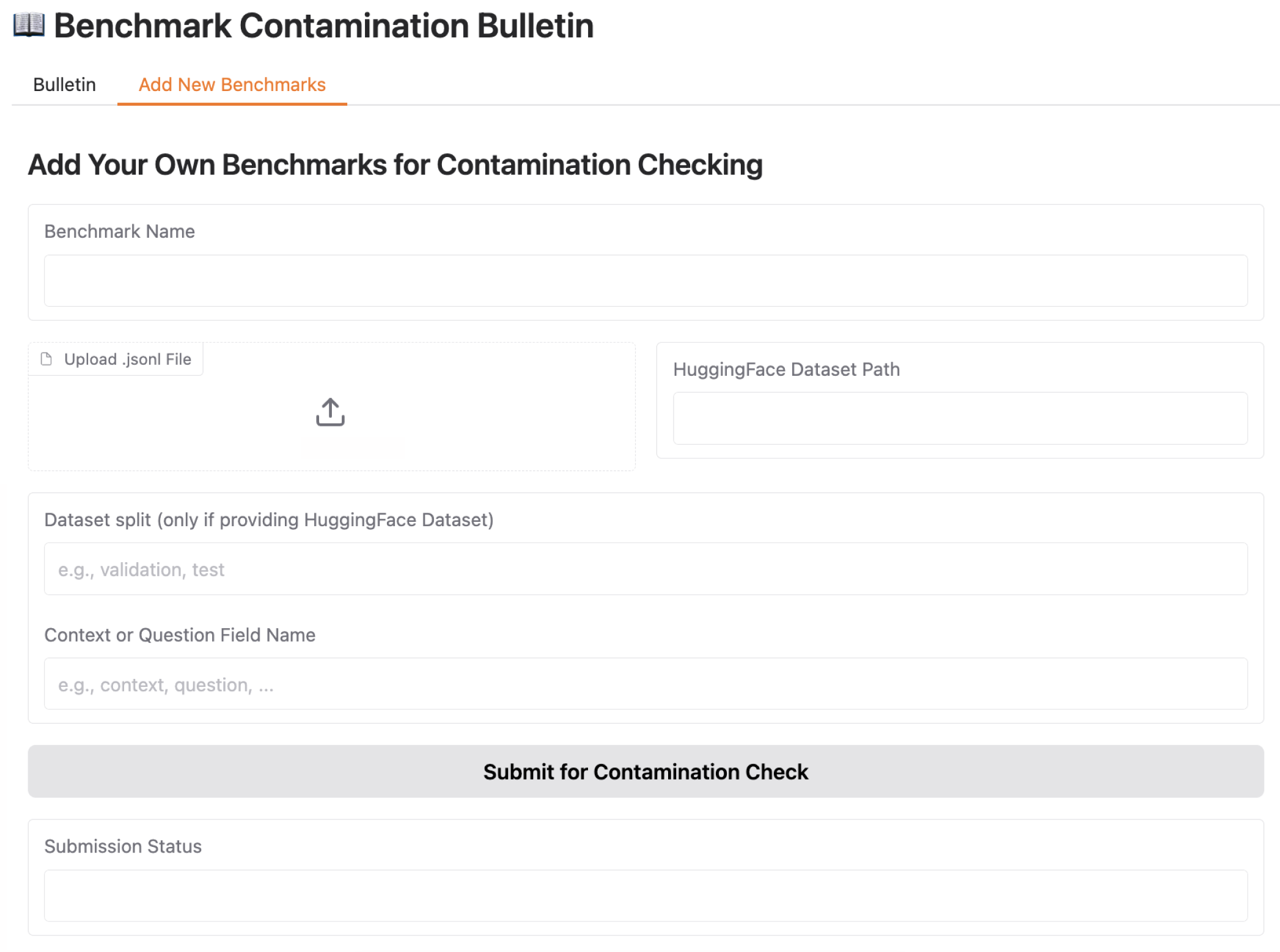}
    \caption{
    \textbf{Upper:} A screenshot of our online Benchmark Contamination Bulletin. 
    \textbf{Lower:} We invite the community to add new benchmarks or upload new ones for contamination analysis, which will be added to the bulletin.
    }
    \label{fig:bulletin_interface}
\end{figure*}

\autoref{fig:bulletin_interface} shows the interface of the system. It consists of two tabs: (1) Benchmark Contamination Bulletin, and (2) submission page for community to contribute. Benchmarks analyzed in this paper are reported in ``core'' table, and submitted benchmarks will be added to ``community'' table.

\section{License of Corpora Used in the Paper}
Pile and DCLM are licensed under the MIT License. Common Crawl is licensed under its customized Limited License. We followed the listed intended use. 

\end{document}